\documentclass[journal]{IEEEtai}

\usepackage[colorlinks,urlcolor=blue,linkcolor=blue,citecolor=blue]{hyperref}

\usepackage{color,array}

\usepackage{graphicx}

\usepackage{amsfonts}
\usepackage{algorithmic}
\usepackage{algorithm}
\usepackage{array}
\usepackage[caption=false,font=normalsize,labelfont=sf,textfont=sf]{subfig}
\usepackage{textcomp}
\usepackage{stfloats}
\usepackage{url}
\usepackage{verbatim}
\usepackage{graphicx}
\usepackage{cite}
\usepackage{times}
\usepackage{helvet}
\usepackage{courier}
\usepackage{amsmath}
\usepackage[normalem]{ulem}
\useunder{\uline}{\ul}{}
\usepackage{float}
\usepackage{makecell}
\usepackage[square, comma, sort&compress, numbers]{natbib}

\usepackage[table,xcdraw]{xcolor}
\usepackage{multirow}
\usepackage{bbding}
\usepackage{booktabs}
\usepackage{amssymb}

\definecolor{green}{rgb}{0.0, 0.5, 0.0}

\definecolor{green2}{RGB}{77, 175, 74}
\definecolor{red2}{RGB}{220, 20, 60}

\usepackage{fancyhdr}
\begin{document}

\title{Integrating Large Language Model for \\ Improved Causal Discovery}

\author{Taiyu Ban, Lyuzhou Chen, Derui Lyu, Xiangyu Wang$^*$, Qinrui Zhu, Qiang Tu, Huanhuan Chen$^*$, ~\IEEEmembership{Fellow,~IEEE}

\thanks{This research was supported in part by the National Key R\&D Program of China (No. 2021ZD0111700), in part by the National Nature Science Foundation of China (No. 62137002, 62176245, 62406302), in part by the  Natural Science Foundation of Anhui province (No. 2408085QF195), in part by the Fundamental Research Funds for the Central Universities under Grant WK2150110035.}
\thanks{Taiyu Ban, Lyuzhou Chen, Derui Lyu, Xiangyu Wang, Qinrui Zhu and Huanhuan Chen are with the School of Computer Science and Technology, University of Science and Technology of China, Hefei 230026, China. Qinrui Zhu is also with Laboratory for Big Data and Decision, University of Science and Technology of China. Qiang Tu is with First Affiliated Hospital of University of Science and Technology of China, Hefei 230026, China.}

\thanks{$^*$\textit{Corresponding authors}: Xiangyu Wang (sa312@ustc.edu.cn) and Huanhuan Chen (hchen@ustc.edu.cn).}

}


\maketitle

\thispagestyle{fancy}

\begin{abstract}
Recovering the structure of causal graphical models from observational data is an essential yet challenging task for causal discovery in scientific scenarios.
Domain-specific causal discovery usually relies on expert validation or prior analysis to improve the reliability of recovered causality, which is yet limited by the scarcity of expert resources.
Recently, Large Language Models (LLM) have been used for causal analysis across various domain-specific scenarios, suggesting its potential as autonomous expert roles in guiding data-based structure learning.
However, integrating LLMs into causal discovery faces challenges due to inaccuracies in LLM-based reasoning on revealing the actual causal structure.
To address this challenge, we propose an error-tolerant LLM-driven causal discovery framework.
The error-tolerant mechanism is designed three-fold with sufficient consideration on potential inaccuracies.
In the LLM-based reasoning process, an accuracy-oriented prompting strategy restricts causal analysis to a reliable range.
Next, a knowledge-to-structure transition aligns LLM-derived causal statements with structural causal interactions.
In the structure learning process, the goodness-of-fit to data and adherence to LLM-derived priors are balanced to further address prior inaccuracies.
Evaluation of eight real-world causal structures demonstrates the efficacy of our LLM-driven approach in improving data-based causal discovery, along with its robustness to inaccurate LLM-derived priors.
Codes are available at \url{https://github.com/tyMadara/LLM-CD}.
\end{abstract}

\begin{IEEEImpStatement}
Structure learning of causal graphical models from observational data is essential for uncovering causal relationships in scientific research. However, recovering true causal structures is challenging due to data insufficiency and noise in real-world scenarios. To address this, our paper integrates Large Language Models (LLMs) to derive variable metadata and apply metadata-driven causality to enhance data-based structure learning. This approach significantly improves the accuracy of recovered causal structures across eight diverse domain-specific datasets, ready to support reliable causal discovery in various research domains. 
It offers a novel expert-free alternative to traditional data-based methods, paving the way for low-cost, high-quality causal discovery systems in various fields.

\end{IEEEImpStatement}


\section{Introduction}

\IEEEPARstart{L}{earning} directed acyclic graph (DAG) structures from observational data is essential for analyzing and discovering causal mechanisms in scientific research \cite{sachs2005causal,kyono2021exploiting,mascaro2023modeling,yang2022online,chen2022directed,moe2021increased}.
This task, known as causal structure learning (CSL) \cite{scutari2019learns,xiang2023bootstrap,vowels2022d,heinze2018causal}, is NP-hard due to the exponentially growing hypothesis space with the number of nodes \cite{chickering1996learning}.
Besides, the limitations of real-world data, such as its insufficiency and noise, further challenge the practical application of structure learning algorithms for domain-specific causal discovery \cite{heckerman1995learning2,de2011efficient}.

Integrating expert knowledge to validate learned structures or provide prior constraints is an effective strategy for ensuring the quality of recovered causal structures \cite{amirkhani2016exploiting,constantinou2023impact,maydeu2020estimating}. However, this approach is limited by the scarcity of expert resources available to verify data-based structure learning seamlessly.
Recently, LLMs, as autonomous cross-domain experts, have shown promise in causal analysis across various domain-specific cases \cite{zhang2023causality,kiciman2023causal,nori2023capabilities,long2023can}. This indicates LLM's potential as substitutes for human experts to address the scarcity of expertise in domain-specific causal discovery.

However, LLM's recognition of causality is limited to the semantic aspect and existing knowledge, challenging its utility in causal discovery. Firstly, the investigated causal mechanisms underlying observational data can remain mysterious to humans. In this context, LLMs can struggle to justify causal relationships between variables with unclear causal mechanisms \cite{zevcevic2023causal,lu2023emergent}. This limitation indicates the unreliability of LLM-based causal reasoning on all pairs of variables, which is a dominant approach in current studies of LLM-based causal analysis \cite{kiciman2023causal,tu2023causal}.
Moreover, graphical models regard causality as a direct interaction or influence implied by data, a view of causality that is not consistent with LLM's recognition, which can be qualitative and ambiguous. Many commonsense causalities derived by LLMs may actually correspond to distinct causal mechanisms, such as direct interactions, multi-step interactions, or confounding by other variables. 
This inconsistency can compromise the recovered structure if not properly addressed in the structure learning process.

To address these challenges, this paper proposes an LLM-driven structure learning framework incorporating error-tolerant mechanisms to mitigate the inaccuracies of LLM-based causal reasoning in data-based structure learning.
The error-tolerant mechanisms are designed in three-fold: 1) Restriction of LLM-based reasoning to clear causal knowledge; 2) Employment of LLM-derived prior as indirect causal interactions; 3) Soft application of prior constraints in structure learning.
Next, we delve into the details of these mechanisms.

For LLM-based reasoning, an accuracy-oriented prompting strategy is designed to address the inaccuracies of LLM-based reasoning on unclear causal mechanisms. This strategy includes three stages: metadata derivation, causal extraction, and causal validation.
The LLM begins by inferring metadata that precisely describes variables based on their semantic symbols and value sets.
Using these metadata, the LLM is prompted to extract causality by restricted single-step reasoning, focusing on clear causal knowledge.
The extracted causality undergoes a chain-of-thought (CoT) \cite{wei2022chain} reasoning and validation process to filter out errors.
Compared to previous pairwise causal reasoning strategies \cite{kiciman2023causal}, this approach avoids forcing the LLM to derive answers about causal relationships between variables with unclear causal mechanisms, thereby ensuring knowledge reliability. Additionally, it requires \(O(1)\) complexity in prompting rather than \(O(n^2)\) when analyzing all variable pairs, leading to significantly better efficiency.

\begin{figure*}[!t]
    \centering
    \includegraphics[width=0.93\textwidth]{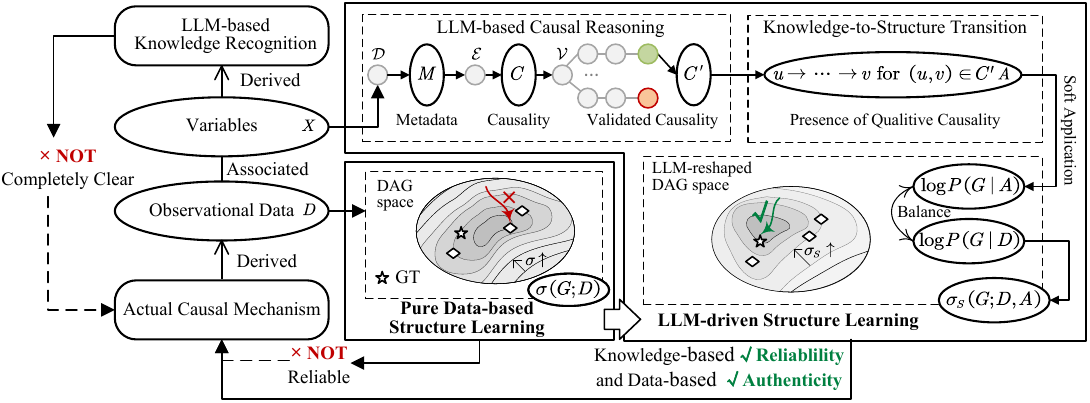}
    \caption{An illustrative diagram of our LLM-driven causal discovery framework. 
    Elliptical blocks represent the input or intermediate results.
    Rounded rectangular blocks represent two aspects of causality among the investigated variables, the actual causal mechanisms (structure) underlying data, and the knowledge recognition of causality among variables.
    Rectangular blocks with the solid border represent a specific process.
    Rectangular blocks with the dotted border represent modules or sub-processes.
    Circles with colored faces in the block `LLM-based Causal Reasoning' represent the reasoning traces of the LLM.
    }
    \label{fig_framework}
\end{figure*}

To align the LLM-derived knowledge with the causal structure, each causal relationship is employed as a positive ancestral constraint where a directed path (or edge) from the cause to effect node exists in the structure.
We do not directly specify the existence of edges as the LLM-derived causal knowledge are essentially qualitative and can represent indirect interactions, represented by paths rather than edges.
This is a critical distinction of this study from previous studies that use LLM to identify edges for causal graph construction.
By this transition, the causal structure specified by LLM-derived knowledge is aligned with its qualitative nature.

In the structure learning process, we employ a soft approach to apply LLM-derived prior constraints to filter out the remaining errors after the preceding two steps.
These errors include misunderstandings due to insufficient metadata or seemingly reasonable statements that actually involve non-causal mechanisms\footnote{Here, the non-causal mechanism means that the interaction between two variables is not represented by a directed path or edge in the causal structure.}. 
Concretely, the scoring criteria for goodness-of-fit to data is extended with a term that evaluates the degree of adherence to prior structures. It creates a balance between goodness-of-fit to data and prior adherence, ensuring that any structural constraint that severely compromises the graphical model's ability to represent the data is not accepted. This mechanism addresses the inconsistency between causality derived from the metadata and the actual causality underlying the observational data, thus improving robustness.

Experimental results on eight real-world datasets spanning diverse domains demonstrate the effectiveness of the proposed LLM-driven causal discovery framework in improving recovered causal structures.
An illustrative diagram is provided in Fig. \ref{fig_framework}.
The main contributions are summarized as follows:
\begin{itemize}
    \item This paper proposes a novel LLM-driven causal discovery framework, which shows a promise in using LLM to aid discovering new causal mechanisms.
    \item The paper adapts LLM-based reasoning to the causal discovery context by limiting the reasoning range to clear, knowledge-based causal relationships, ensuring the quality of LLM-derived priors. 
    \item The paper aligns LLM-derived causality with its qualitative nature as ancestral constraints on the causal structure and applies them softly. This knowledge-to-structure transition and balancing ensures the stability of recovered causality by LLM-driven structure learning.
\end{itemize}

The rest of the paper is organized as follows.
Section \ref{sec:related_work} introduces the background of this study.
Section \ref{sec:method} introduces the proposed LLM-driven CSL method.
Section \ref{sec:experiments} reports the experimental results and performs the analysis.
Section \ref{sec:conclusions} concludes this paper.

\section{Background}
\label{sec:related_work}

This section reviews related studies on the application of LLM-based causal reasoning in causal tasks, followed by preliminaries of structure learning in causal discovery.

\subsection{LLM-based Reasoning in Causal Analysis}
The research interest in exploring LLM's reasoning ability in causal analysis has emerged recently \cite{zhang2023understanding}.
In the context of LLM performance measurement, Jin \textit{et al.} \cite{jin2023can} propose a novel task that evaluates LLM's ability in extracting causality among a set of correlation statements.
Similarly, Willig \textit{et al.} \cite{willig2022can} investigate LLM's proficiency in responding to causality-based queries, assessing their capability to discern causality and correlation. 
In domian-specific causal analysis contexts, Long \textit{et al.} \cite{long2023can} employ LLM to generate simple causal structures for small variable sets (3-4 variables), examining LLM's accuracy in determining causal directions between pairs of variables.
Another study by Tu \textit{et al.} \cite{tu2023causal}, challenges LLM with the task of discovering causal structures within the medical pain diagnosis domain. 
Their findings indicate suboptimal performance of LLM in this task. 
For the same dataset, Kiciman \textit{et al.} \cite{kiciman2023causal} obtain more refined results with LLM by optimizing the prompting strategies. 
Moreover, they extend their investigation to a wider array of causal tasks, including pairwise causal discovery \cite{hoyer2008nonlinear} and counterfactual reasoning \cite{frohberg2021crass}, where LLM demonstrate notable accuracy.

In addition to these promises, there are some doubts about LLM's genuine understanding of causality. Zečević \textit{et al.} \cite{zevcevic2023causal} present empirical analyses questioning the ability of LLM to reason about causality, suggesting that they do not exhibit authentic causal understanding. 
Lu \textit{et al.} \cite{lu2023emergent} raise doubts about whether LLM's apparent reasoning abilities are merely the result of in-context learning and provide experimental evidences across various tasks.
These findings indicate that LLM alone can fail to discover new causal relationships in domain-specific cases.

Unlike previous studies that use LLM as standalone systems for causal analysis, we deploy LLM as supplementary tools in causal discovery. This integration compensates for the LLM's limitations in revealing new causality by incorporating observational data, while simultaneously enhancing data-based analysis with the guidance of the LLM's causal insights.

\subsection{Use of LLM Prior in Data-based Analysis}
The closest studies to this work are those by Choi \textit{et al.} \cite{choi2022lmpriors} and Long \textit{et al.} \cite{long2023can}, which incorporate LLM-derived priors into the process of data-based analysis. Choi \textit{et al.} \cite{choi2022lmpriors} explore the use of LLM-based reasoning results as priors in three data-based analysis tasks, including causal directionality inference between pairwise variables, and have observed improved results. Long \textit{et al.} \cite{long2023can} integrate LLM-reasoned directionality results of edges between pairwise variables using a greedy strategy to orient edges in an undirected causal graph.

A major difference between these studies and this paper is our investigation into the potential of LLM for a more complex and applicable task: data-based structure learning for causal discovery. Unlike the study by Long \textit{et al.}, which operates on a pre-existing skeleton of causality, our approach directly recovers structures from observational data.
Moreover, the reliability of LLM-based causal reasoning on pairwise variables is significantly lower in our context, as causal discovery involves a large number of variable pairs with unclear causal mechanisms. This complexity presents unique challenges that our framework aims to address.

\subsection{Causal Structure Learning}
We employ the Bayesian network (BN) \cite{pearl2009causality} as the causal graphical model\footnote{In additional to BNs, the structural equation model (SEM)  is also a well-known causal graphical model, which typically models the causality among continuous variables as functions with additive noise \cite{zheng2018dags,yang2024additive}.}, which represents the conditional dependencies among nodes \(X = \{x_i\}_{i=1}^n\) via a directed acyclic graph (DAG).
The joint probability distribution governed by the BN is defined as follows:
\begin{equation}
    P(x_1,x_2,...,x_n) = \prod_{i=1}^{n} P(x_i\mid \text{pa}(x_i))
    \label{eq:markov}
\end{equation}
where $\text{pa}(x_i)$ is the set of parent nodes of $x_i$ in the BN.
Structure Learning of a BN is to deduce its DAG structure from a set of i.i.d samples generated accordingly \cite{de2009structure}.

Methods for BN structure learning are typically categorized into constraint-based, score-based, and hybrid approaches \cite{vowels2022d}. Constraint-based methods utilize conditional independence tests to establish the graph, resulting in a partial DAG \cite{scutari2017bayesian}. Score-based methods evaluate the graph’s goodness-of-fit to the data using a scoring criteria and search for the DAG with the optimal score \cite{bouchaala2010improving}. Hybrid methods are a combination of these approaches \cite{gasse2012experimental}.

In this paper, we use score-based methods as structure learning algorithms given that their extensibility to prior structural constraints \cite{constantinou2023impact,li2018bayesian}. Given the Markov property of joint data distributions, as shown in Eq. \eqref{eq:markov}, the mainstream scoring functions like Bayesian Information Criterion (BIC) \cite{schwarz1978estimating}, Minimum Description Length (MDL) \cite{rissanen1978modeling}, and Bayesian Dirichlet equivalent uniform (BDeu) \cite{buntine1991theory}, are decomposable \cite{heckerman1995learning}.
They can be presented as the following form:
\begin{equation}
\sigma(G;D)=\sum_{i=1}^n \sigma \left(x_i, \operatorname{pa}\left(x_i\right);D\right)
\label{eq:decomp}
\end{equation}
where \(G\) denotes the BN structure and \(D\) represents the observational data.
Score-based structure learning is framed as a combinatorial optimization task aimed at maximizing \(\sigma(G; D)\), while ensuring that \(G\) remains acyclic.
Representative search strategies in score-based structure learning include Greedy Equivalent Search \cite{chickering2002optimal}, Tabu Search \cite{li2018bayesian}, A* \cite{yuan2011learning}, and Markov chain Monte Carlo \cite{o2006learning} algorithms, etc.

\section{LLM-driven Causal Structure Learning}
\label{sec:method}

This section illustrates the proposed LLM-driven structure learning approach. We begin by introducing the LLM-based causal reasoning among variables with a focus on accuracy prioritization. Then, we describe the LLM-derived prior transition to structural constraints and apply them softly in the scoring criteria of graphs. Next, we present the search algorithm used to find the optimal LLM-guided causal structure. Finally, we discuss the error-tolerance ability and introduce an alternative hard approach for applying LLM-derived priors.

To set the stage, we assume that the variable set is $X=\{x_i\}_{i=1}^n$, and the set of observational data is discrete, denoted as $D\in \mathbb{N}^{m\times n}$.
A directed graph with these variables is denoted as $G(X,E(G))$, where $X$ is the node set, and $E(G)\subseteq X\times X$ is the edge set of this graph.
The scoring criteria of the goodness-of-fit to data of a graph is denoted as $\sigma(G;D)$, which is decomposable as defined in Eq. \eqref{eq:decomp}.

\begin{figure}
    \centering
    \includegraphics[width=.48\textwidth]{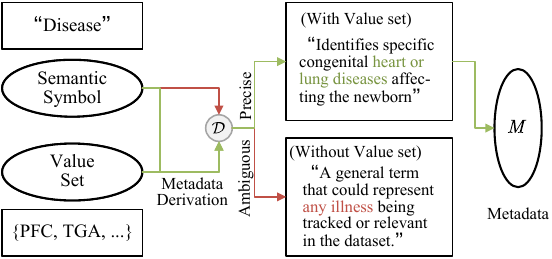}
    \caption{An illustrative diagram of the metadata derivation stage of the prompting strategy. The example in the figure is selected from outputs of GPT-4 on the Child dataset, where integration of semantic symbols and value sets leads to more precise and contextually accurate metadata.}
    \label{fig:meta_der}
\end{figure}

\subsection{Accuracy-Oriented Causal Reasoning with LLM}

This section introduces the accuracy-oriented prompting strategy for LLM-based causal reasoning among variables.
The LLM-based reasoning consists of three stages, which are the metadata derivation, causal extraction, and causal validation.

Assume that the observational data of causal discovery task contains basic descriptive information, including variables' semantic symbols, value sets and the overall research context.
The details of the prompting strategy are illustrated as follows.

In the metadata derivation $\mathcal{D}$ stage, the LLM is prompted to identify richer metadata on the variables given their symbols, value sets, and context. Note that the value set is crucial for enhancing the precision of metadata. Some variable symbols represent general concepts and can be ambiguous; integrating possible values helps the LLM provide more precise descriptions that align with their reference in the research context. Representative examples are reported in Table \ref{tab:metadata}, comparing the differences between derived metadata using only symbols and those using both symbols and value sets. The derived metadata for variables is denoted as \( M = \{ \text{meta}_i \}_{i=1}^n \), with \( \text{meta}_i \) as descriptive texts for variable \( x_i \).
An illustrative diagram of the metadata derivation is presented in Fig. \ref{fig:meta_der}.

In the causal extraction \(\mathcal{E}\) stage, the LLM is prompted to extract the most confident causal relationships among the variables \(\{x_i\}_{i=1}^n\) given their metadata \(\{ \text{meta}_i \}_{i=1}^n\). We avoid providing pairwise variables individually to the LLM, as this can lead to unreliable causality extraction for variable pairs whose interaction mechanisms are unclear. Studies have shown that LLM can perform more fine-grained reasoning on complex tasks when decomposed into sub-tasks \cite{zhao2023enhancing,zhou2022least}. Conversely, by restricting the LLM to single-step reasoning, we focus on the relatively easier part of the causal extraction task, allowing the LLM to output the most clear causality in its knowledge to prioritize accuracy. The set of extracted causality is denoted as \(C=\{(u,v) \mid u,v \in X\}\).

\begin{figure*}[!t]
    \centering
    \includegraphics[width=0.9\textwidth]{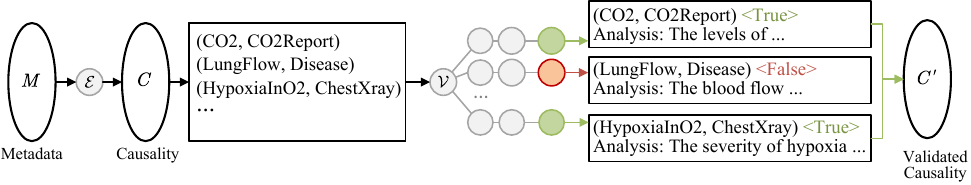}
    \caption{An illustrative diagram of the causal extraction and validation stages. The LLM is first prompted to extract causality among variables using their metadata, followed by a decomposition and validation process to ensure the correctness of each extracted causal statement.}
    \label{fig:causal_ext_val}
\end{figure*}

In the causal validation \(\mathcal{V}\) stage, the extracted causality set \(C\) is validated through decomposition and CoT prompting. This validation stage is necessary because the LLM's reasoning ability is limited in complex tasks, which can lead to erroneous results. To address this inaccuracy, we have the LLM decompose each causal relationship in \(C\) and conduct a CoT reasoning to validate its correctness. By this stage, errors in the extracted causality are filtered out to ensure the accuracy of the prior causality that guides data-based structure learning. The validated causality is denoted as \(C^\prime\).
An illustrative diagram of causal extraction and validation stages is shown in Fig. \ref{fig:causal_ext_val}.

\begin{table*}[!t]
\centering
\caption{Comparative examples of LLM-derived metadata using only symbols versus using both symbols and value sets. }
\begin{tabular}{@{}llll@{}}
\toprule
Symbol & Value set                     & Metadata with Both Symbol and Value Set                                     & Metadata with Only Symbol                \\ \midrule
BirthAsphyxia   & \{yes, no\}                   & Indicates if birth complications led to \textcolor{green}{oxygen deprivation}                  & ... suffered from \textcolor{green}{lack of oxygen} ...     \\
HypDistrib      & \{Equal, Unequal\}            & Shows the \textcolor{green}{balance of oxygen} deficiency across tissues                       & Describes the \textcolor{red}{distribution} pattern of .. \\
Disease         & \{PFC, TGA, ...\}             & Identifies specific congenital \textcolor{green}{heart or lung} diseases ... & Represents \textcolor{red}{any specific} illness ...      \\
LungFlow        & \{Normal, Low, High\}         & Measures the amount of \textcolor{green}{blood flow} through the lungs                         & ... \textcolor{red}{air or blood} ...                     \\
LowerBodyO2     & \{\textless{}5, 5-12, 12+\}   & Measures peripheral \textcolor{green}{oxygen saturation} ...  & ... \textcolor{green}{oxygen saturation} ...       \\
XrayReport      & \{Oligaemic, Plethoric, ...\} & Provides a summary of X-ray findings on \textcolor{green}{lung health} ... & ... \textcolor{red}{bone structures, organs} ...          \\ \bottomrule
\end{tabular}
\begin{flushleft}
These examples are generated by GPT-4 on the \textit{Child} dataset. Red text indicates inaccurate descriptions of variables in the overall research context of the \textit{Child} dataset, while green text indicates precise descriptions.
\end{flushleft}
\label{tab:metadata}
\end{table*}

\subsection{Soft Scoring Criteria with LLM-derived Prior}
\label{sec_method_anc}

This section introduces how to softly extend the scoring criteria to incorporate LLM-derived priors.
We begin with the structural constraints of the LLM-derived causality in $C^\prime$, which is converted to a set of positive ancestral constraints on the causal graph $G$:
\begin{equation}
    A = \{\text{Directed path } u\leadsto v \text{ is in }G \text{ for } (u,v)\in C^\prime\}
\end{equation}
To softly incorporate $A$ in structure learning, we require a prior probability set $\mathcal{P}$ to construct the prior \(\Pi\langle A,\mathcal{P} \rangle\).
Each constraint \(x_i\leadsto x_j\) within \(A\) is associated with a prior probability \(\mathcal{P}_{ij}\), constrained within the range \((0.5,1)\) to indicate the presence of the path.
Given observational data \(D\) and the prior knowledge of ancestral constraints \(\Pi\), the posterior probability of a DAG \(G\) is formulated as follows:
\begin{equation}
\begin{split}
    P(G \mid D,\Pi) = \frac{P(D \mid G)P(G\mid \Pi)}{P(D\mid \Pi)}
\end{split}
\label{eq:bayes}
\end{equation}
Eliminating the constant \(P(D\mid \Pi)\), the typical form of scoring functions is derived by the logarithmic transformation of Eq. \eqref{eq:bayes}:
\begin{equation}
\log P(G\mid D,\Pi) = \log P(D\mid G) + \log P(G\mid \Pi)
\label{eq:common}
\end{equation}
Here, \(\log P(D\mid G)\) assesses the goodness-of-fit of the graph to data. 
The term \(\log P(G\mid \Pi)\) evaluates the adherence to prior constraints of the graph, which is discussed subsequently.

For the set of ancestral constraints $A$, the existence of paths typically involves interdependencies, leading to complex joint distribution forms. Nevertheless, given that the ancestral constraints specified are generally sparse, we can approximate the probability distribution by calculating the existence of each path independently. This approach simplifies the computation of the score while maintaining the ancestral constraints' appropriate impact on the data-consistency term within the scoring criteria. This factorization is expressed as follows:
\begin{equation}
\begin{split}
    &\log P(G\mid \Pi) = c_0 + \\
    &\sum_{x_i\leadsto x_j \in A} \log \left( \mathbb{I}_{x_i\leadsto x_j\in G} \mathcal{P}_{ij} + \mathbb{I}_{x_i \leadsto x_j\not\in G}(1-\mathcal{P}_{ij}) \right)
\end{split}
\label{eq:prior_term}
\end{equation}
Here, \(c_0\) is a constant, and \(\mathbb{I}_{\texttt{condition}}\) is an indicator function that returns 1 if \texttt{condition} is true and 0 otherwise.
Combining Equations \eqref{eq:common} and \eqref{eq:prior_term} and dropping constants, the ancestral constraint-based scoring function is formulated as:
\begin{equation}
\begin{split}
&\sigma_S(G;D,\Pi) = \sigma(G;D) + \\
    &\sum_{x_i\leadsto x_j \in A} \log \left( \mathbb{I}_{x_i\leadsto x_j\in G} \mathcal{P}_{ij} + \mathbb{I}_{x_i \leadsto x_j\not\in G}(1-\mathcal{P}_{ij}) \right)
\end{split}
\label{eq:prior_score}
\end{equation}
In this equation, $\sigma_S$ denotes the soft constraint scoring function, integrating ancestral constraints $A$ with the prior probability set $\mathcal{P}$. The term \(\sigma(G;D)\) is the scoring function that estimates \(\log P(D\mid G)\).

The soft constraint scoring function in Eq. \eqref{eq:prior_score} is used as the optimization objective during the DAG search process. In practice, the specific prior probability of each individual constraint is often unknown, and the prior probability $\mathcal{P}$ is assigned a uniform value that represents the overall confidence of the ancestral constraints, denoted as $\mathcal{P}_{ij}\gets c$.

This confidence level has a significant impact on the search results, as it affects the impact strength of prior constraints on the search process. A higher confidence level will lead to a greater tendency to accept constraints, while a lower confidence level will result in a greater willingness to consider alternative solutions that may better fit the data.

The choice of confidence level is problem-dependent and requires careful consideration. A too-high confidence level may lead to an excessive reliance on prior knowledge, potentially overlooking important aspects of the data. Conversely, a too-low confidence level may result in an lack of consideration of prior knowledge, leading to suboptimal solutions.

In Section \ref{sec:confidence}, we will discuss the impact of confidence levels on the search results and how to choose an appropriate confidence level for a given problem.

\subsection{Search with LLM-driven Scoring Criteria}
\label{sec_method_soft_search}

To search for the optimal structure with LLM-driven scoring criteria, we employ an ordering-based search strategy that assumes a total ordering of variables $\prec_o$.
Within this ordering, the acyclicity naturally holds by assigning the candidate parent set of each variable as variables preceding it in $\prec_o$.
The search strategy consists of two paradigms: one for search in the graph space within total ordering $\prec_o$, and another for search in the order space. We use the MCMC method proposed by O'Donnell et al. \cite{o2006learning} for implementation.

The data-driven scoring function $\sigma(G;D)$ is extended to the LLM-driven scoring function $\sigma_S(G;D,\Pi)$ as illustrated earlier. For the two search paradigms, the mappings of a sample to its neighbor are defined as follows:
\begin{enumerate}
    \item The neighbor of an ordering $\prec_o$ ($\mathcal{N}_{\prec_o}$): swap the position of two random variables in $\prec_o$.
    \item The neighbor of a DAG $G$ ($\mathcal{N}_{G}$): With adherence to the ordering, toggle a single edge between arbitrary two nodes, toggle two edges from arbitrary two nodes to one, and then swap the parent node of one node to another.
\end{enumerate}
With the above neighborhood definition, an MCMC search strategy is employed, starting with an initial best ordering $\prec_o^0$ and its corresponding DAG $G^0$ derived by Simulated Annealing. Then, a random neighbor of the initial ordering is sampled, denoted as $\mathcal{N}_{\prec_o^0}$, followed by sampling a random neighbor of the initial DAG $\mathcal{N}_{G^0}$ with adherence to the new ordering $\mathcal{N}_{\prec_o^0}$.
The new ordering sample is accepted with the following condition:
\begin{equation}
   \frac{\sigma_S({\mathcal{N}_{G^0};D,\Pi}) - \sigma_S({G^0;D,\Pi})}{T} > \log U[0,1] 
\end{equation}
where $T$ is the temperature\footnote{$T$ is set to $1.8$ as default.} and $U[0,1]$ is a uniform random number from $0$ to $1$.
If $\mathcal{N}_{\prec_o^0}$ is accepted, a weight \(\frac{\exp(\sigma_S({\mathcal{N}_{G^0};D,\Pi})-\exp(\sigma_0)}{T}\) is contributed to the `clean' DAG of $N_{G_0}$, where $\sigma_0$ is a constant to avoid underflow.
The `clean' DAG is derived by eliminating any single edge of $G_0$ that can refine the score \(\sigma_S\).
Then the new ordering $\mathcal{N}_{\prec_o^0}$ and DAG $\mathcal{N}_{G^0}$ are taken as the input to the next iteration of samples until a set number of samples are complete.
When the sampling process is over, the clean DAGs are grouped using a Kullback-Leibler (KL) divergence test, and the DAG with highest weight is taken as the solution.

\subsection{Impact of Confidence Level}
\label{sec:confidence}

The confidence level, denoted by the prior probability $c$, is pivotal in shaping the error tolerance of the algorithm. It directly affects the penalty imposed for violating ancestral constraints, as detailed in Eq. \eqref{eq:prior_score}. As the confidence level nears unity, the associated penalty value escalates sharply.

To elucidate the effect of the confidence level on penalty severity, consider $c = 1 - 10^{-n}$, where $n$ belongs to the set of positive integers, $\mathbb{N}^+$. This assumption enables the formulation of the penalty function as follows:
\begin{equation}
\log \frac{c}{1-c} = \log(1-10^{-n}) + n \log 10 \approx n \log 10,
\end{equation}
where the approximation is valid for larger values of $n$. This demonstrates that the penalty value increases linearly with $n$, while the confidence level is taken as $1 - 10^{-n}$.

The correlation between the confidence level and penalty strength underscores a crucial point: if the algorithm accepts fewer prior constraints than anticipated, adjusting the confidence level to $1-10^{-n}$ enables the modification of constraint strength through alterations in $n$. This adjustment facilitates the calibration of the algorithm’s tolerance to errors, allowing it to be tailored to the specific demands of the task.

\subsection{A Hard Alternative of LLM-driven Structure Learning}
\label{sec_method_hard}

This paper also investigates the hard approach of applying LLM-derived ancestral constraints in the CSL process.
This approach strictly adheres to all constraints, provided that they are not in conflict. 
It can be regarded as an equivalent operation to setting the prior probability of all prior constraints to 1.
Considering the prior probability-based scoring function in Eq. \eqref{eq:prior_score}, we set $P_{ij}=1$, which makes the violation to a constraint result in a negative infinity penalty to the score ($\lim_{c\to 0}\log c -\log 1$). Therefore, the hard constraint scoring function can be factorized as follows:
\begin{equation}
\sigma_H(G;D,A) = \sigma(G;D) - \inf \times \psi(G;A)
\end{equation}
where \(\psi(G;A)\) represents the count of ancestral constraints that are not satisfied in the graph \(G\), and \(\inf\) is a positive value large enough to ensure that all graphs fully satisfying \(A\) score higher than those that do not.
To find the solution with the optimal hard prior constraint scoring function \(\sigma_H\), we use an order-based search proposed by Lee \textit{et al.} \cite{lee2017metaheuristics}. This method employs a Memetic search algorithm with a tabu list and an Insert Neighbourhood definition in the order space.

\begin{table}[!h]
\centering
\caption{The used casual structure datasets.}
\small
\setlength{\tabcolsep}{4pt}
\begin{tabular}{@{}c|c|c|c|c@{}}
\hline
Dataset   & \# V & \# E & \# Para &  Domain of Research                         \\ \hline
Cancer    & 5         & 4     & 10             & Smoking and cancer             \\
Asia      & 8         & 8     & 18             & Reasons causing dyspnoea       \\
Child     & 20        & 25    & 230           & Childhood diseases             \\
Alarm     & 37        & 46    & 509          & ALARM monitoring system        \\
Insurance & 27        & 52    & 1008          & Car insurance cost             \\
Water	& 32 & 66 &10k	&	Waste water control systems \\
Mildew    & 35        & 46    & 540k        & Winter wheat mildew control    \\
Barley    & 48        & 84    & 114k       & Pesticides use in growing malt \\ 
\hline
\end{tabular}
\begin{flushleft}
    \footnotesize Column headers `\# V', `\# E', and `\# Para' represent the number of variables, edges, and parameters, respectively.
\end{flushleft} 
\label{tab_datasets}
\end{table}

\begin{table}[!t]
\centering
\caption{Comparison between CSL with mere data and with LLM-derived prior by the soft constraint approach (MML score).}
\begin{tabular}{c|c||lll}
\multirow{2}{*}{Dataset}   & \multirow{2}{*}{N} & \multicolumn{3}{c}{F1$\uparrow$ / SHD$\downarrow$}                           \\
                           &                    & Data-based  & LLM-driven  & Improvement                                      \\ \hline
\multirow{2}{*}{Cancer}    & 250                & 0.67 / 2.0  & 0.75 / 2.0  & \textcolor{red}{+12\%} / \textcolor{red}{-0\%}    \\
                           & 1000               & 0.45 / 2.5  & 0.83 / 1.3  & \textcolor{red}{+84\%} / \textcolor{red}{-47\%}  \\ \hline
\multirow{2}{*}{Asia}      & 250                & 0.68 / 3.5  & 0.99 / 0.2  & \textcolor{red}{+46\%} / \textcolor{red}{-95\%}  \\
                           & 1000               & 0.78 / 2.2  & 1.00 / 0.0  & \textcolor{red}{+28\%} / \textcolor{red}{-100\%} \\ \hline
\multirow{2}{*}{Child}     & 500                & 0.83 / 6.0  & 0.82 / 7.0  & \textcolor{blue}{-1\%} / \textcolor{blue}{+17\%}   \\
                           & 2000               & 0.92 / 2.0  & 0.96 / 1.2  & \textcolor{red}{+4\%} / \textcolor{red}{-42\%}     \\ \hline
\multirow{2}{*}{Insurance} & 500                & 0.48 / 34.3 & 0.62 / 28.3 & \textcolor{red}{+29\%} / \textcolor{red}{-17\%}  \\
                           & 2000               & 0.54 / 31.7 & 0.64 / 26.0 & \textcolor{red}{+19\%} / \textcolor{red}{-18\%}  \\ \hline
\multirow{2}{*}{Alarm}     & 1000               & 0.82 / 11.0 & 0.91 / 6.8  & \textcolor{red}{+11\%} / \textcolor{red}{-38\%}  \\
                           & 4000               & 0.87 / 8.2  & 0.97 / 2.5  & \textcolor{red}{+11\%} / \textcolor{red}{-69\%}  \\ \hline
\multirow{2}{*}{Water}     & 1000               & 0.25 / 59.0 & 0.28 / 63.5 & \textcolor{red}{+12\%} / \textcolor{blue}{+8\%}   \\
                           & 4000               & 0.33 / 53.2 & 0.37 / 55.0 & \textcolor{red}{+12\%} / \textcolor{blue}{+3\%}   \\ \hline
\multirow{2}{*}{Mildew}    & 8000               & 0.28 / 48.2 & 0.18 / 55.2 & \textcolor{blue}{-36\%} / \textcolor{blue}{+15\%}  \\
                           & 32000              & 0.21 / 62.2 & 0.20 / 62.3 & \textcolor{blue}{-5\%} / \textcolor{blue}{+0\%}    \\ \hline
\multirow{2}{*}{Barley}    & 2000               & 0.29 / 81.5 & 0.28 / 85.7 & \textcolor{blue}{-3\%} / \textcolor{blue}{+5\%}  \\
                           & 8000               & 0.33 / 81.2 & 0.34 / 81.7 & \textcolor{red}{+3\%} / \textcolor{blue}{+1\%}  
\end{tabular}
\label{tab_cmp_soft}
\end{table}

\begin{table}[!h]
\centering
\caption{Comparison between CSL with mere data and with LLM-derived prior by the hard constraint approach (BDeu score).}
\begin{tabular}{c|c||lll}
\multicolumn{1}{c|}{\multirow{2}{*}{Dataset}} & \multirow{2}{*}{N} & \multicolumn{3}{c}{F1$\uparrow$ / SHD$\downarrow$}                                                  \\
\multicolumn{1}{c|}{}                         & \multicolumn{1}{c||}{}                           & Data-based  & LLM-driven  & Improvement                                       \\ \hline
\multirow{2}{*}{Cancer}                       & 250                                             & 0.58 / 3.0  & 0.94 / 0.5  & \textcolor{red}{+\%62} / \textcolor{red}{-\%83}   \\
                                              & 1000                                            & 0.60 / 1.8  & 1.00 / 0.0  & \textcolor{red}{+\%67} / \textcolor{red}{-\%100}  \\ \hline
\multirow{2}{*}{Asia}                         & 250                                             & 0.68 / 4.2  & 0.88 / 2.0  & \textcolor{red}{+\%29} / \textcolor{red}{-\%48}   \\
                                              & 1000                                            & 0.75 / 2.5  & 0.98 / 0.3  & \textcolor{red}{+\%31} / \textcolor{red}{-\%87}   \\ \hline
\multirow{2}{*}{Child}                        & 500                                             & 0.72 / 9.5  & 0.74 / 10.5 & \textcolor{red}{+\%3} / \textcolor{blue}{+\%11}   \\
                                              & 2000                                            & 0.82 / 5.3  & 0.78 / 7.8  & \textcolor{blue}{-\%5} / \textcolor{blue}{+\%47}  \\ \hline
\multirow{2}{*}{Insurance}                    & 500                                             & 0.68 / 25.7 & 0.70 / 24.5 & \textcolor{red}{+\%3} / \textcolor{red}{-\%5}     \\
                                              & 2000                                            & 0.81 / 15.0 & 0.78 / 16.2 & \textcolor{blue}{-\%4} / \textcolor{blue}{+\%8}   \\ \hline
\multirow{2}{*}{Alarm}                        & 1000                                            & 0.88 / 9.5  & 0.85 / 12.3 & \textcolor{blue}{-\%3} / \textcolor{blue}{+\%30}  \\
                                              & 4000                                            & 0.91 / 6.5  & 0.88 / 8.8  & \textcolor{blue}{-\%3} / \textcolor{blue}{+\%36}  \\ \hline
\multirow{2}{*}{Water}                        & 1000                                            & 0.38 / 62.3 & 0.34 / 66.7 & \textcolor{blue}{-\%11} / \textcolor{blue}{+\%7}  \\
                                              & 4000                                            & 0.48 / 53.7 & 0.47 / 55.7 & \textcolor{blue}{-\%2} / \textcolor{blue}{+\%4}   \\ \hline
\multirow{2}{*}{Mildew}                       & 8000                                            & 0.63 / 22.8 & 0.42 / 40.5 & \textcolor{blue}{-\%33} / \textcolor{blue}{+\%77} \\
                                              & 32000                                           & 0.67 / 21.0 & 0.71 / 21.5 & \textcolor{red}{+\%6} / \textcolor{blue}{+\%2}    \\ \hline
\multirow{2}{*}{Barley}                       & 2000                                            & 0.62 / 47.0 & 0.63 / 52.0 & \textcolor{red}{+\%2} / \textcolor{blue}{+\%11}    \\
                                              & 8000                                            & 0.73 / 33.7 & 0.57 / 54.5 & \textcolor{blue}{-\%22} / \textcolor{blue}{+\%62}
\end{tabular}
\label{tab_cmp_hard}
\end{table}

\section{Experiments}
\label{sec:experiments}
\subsection{Datasets and Setup}

\begin{table*}[!h]
    \centering
    \footnotesize
    \caption{Comparison of SID$\downarrow$ between CSL with mere data and with LLM-derived prior.}
    \setlength{\tabcolsep}{2.5pt}
    \begin{tabular}{cccccccccccccccccccccc}
        \toprule
        \multirow{2}{*}{\makecell{}} & \multirow{2}{*}{Method} & \multicolumn{2}{c}{Cancer} & \multicolumn{2}{c}{Asia} & \multicolumn{2}{c}{Child} & \multicolumn{2}{c}{Insurance} & \multicolumn{2}{c}{Alarm} & \multicolumn{2}{c}{Water} & \multicolumn{2}{c}{Mildew} & \multicolumn{2}{c}{Barley} \\
        \cmidrule(lr){3-4} \cmidrule(lr){5-6} \cmidrule(lr){7-8} \cmidrule(lr){9-10} \cmidrule(lr){11-12} \cmidrule(lr){13-14} \cmidrule(lr){15-16} \cmidrule(lr){17-18}
        & & 250         & 1000         & 250         & 1000       & 500         & 2000        & 500           & 2000          & 1000        & 4000        & 1000        & 4000        & 8000        & 32000        & 2000         & 8000        \\
        \midrule
        \multirow{2}{*}{} 
        & Data-based  MCMC           & 2.00           & 8.00            & 14.5        & 9.67       & 87.0          & 27.5        & 464           & 396        & 119      & 89.3       & 537      & 554      & 693       & 698          & 1235      & 1209        \\
        
        & LLM-driven  MCMC           & 1.00           & 0.67         & 0.00           & 0.00          & 75.2       & 4.00           & 427        & 347         & 47.2       & 26.2       & 520      & 543      & 731      & 683       & 1222         & 1196     \\
        
        \multicolumn{2}{c}{Improvement}  & \textbf{-50\%}    & \textbf{-92\%}     & \textbf{-100\%}   & \textbf{-100\%}  & \textbf{-14\%}    & \textbf{-85\%}    & \textbf{-8.0\%}       & \textbf{-13\%}      & \textbf{-60\%}    & \textbf{-71\% }   & \textbf{-3.2\% }    & \textbf{-1.9\%}     & +5.6\%      & \textbf{-2.1\%}      & \textbf{-1.1\%}      & \textbf{-1.1\%}     \\

        \midrule
        \multirow{2}{*}{} 
        & Data-based Tabu             & 6.25        & 6.83         & 12.2       & 10.7      & 126      & 86.3       & 344        & 260        & 57.3       & 38.7       & 465         & 435      & 426         & 350       & 886      & 730      \\
        
        & LLM-driven  Tabu            & 0.50         & 0.00            & 1.50         & 0.00          & 100      & 82.2       & 342        & 278        & 50.5        & 30.7       & 487       & 454      & 552      & 291       & 837          & 906      \\
        \multicolumn{2}{c}{Improvement} & \textbf{-92\%}    & \textbf{-100\% }   & \textbf{-88\% }   & \textbf{-100\% } & \textbf{-21\%}    & \textbf{-4.8\% }    & \textbf{-0.8\% }      & +6.9\%        & \textbf{-12\% }   & \textbf{-21\%}    & +4.6\%      & +4.3\%      & +30\%     & \textbf{-17\%}     & \textbf{-5.6\%}      & +24\%\\
        \bottomrule
    \end{tabular}
    \label{tab:SID_comp}
\end{table*}

\begin{table*}[!t]
\centering
\small
\setlength{\tabcolsep}{5pt}
\caption{Accuracy of prior ancestral constraints derived from various LLMs on various datasets.}
\begin{tabular}{c|cccccccccccc}
\hline
LLM          & Cancer & Asia   & Child   & Insurance & Alarm   & Water  & Mildew  & Barley  & \textbf{Truth} & \textbf{Error} & \textbf{Accuracy} \\ \hline
Chatsonic & 3 / 5 & 7 / 7 & 6 / 14 & 16 / 24 & 10 / 20 & 0 / 5 & 8 / 24 & 3 / 9 & 53 & 55 & 0.51 \\
HuggingChat & 1 / 4 & 2 / 6 & 2 / 9 & 3 / 5 & 0 / 1 & 0 / 0 & 0 / 0 & 0 / 2 & 8 & 19 & 0.30\\
Whispr & 4 / 5 & 6 / 7 & 1 / 13 & 9 / 14 & 15 / 29 & 2 / 8 & 24 / 30 & 7 / 16 & 68 & 54 & 0.56\\
Xinhuo & 1 / 1 & 1 / 1 & 1 / 1 & 4 / 20 & 0 / 2 & 0 / 0 & 1 / 29 & 1 / 1  & 9 & 46 & 0.16\\
Yiyan & 8 / 15 & 6 / 7 & 4 / 19 & 10 / 26 & 3 / 10 & 0 / 12 & 0 / 18 &  4 / 12 & 35 & 84 & 0.29\\
ChatGLM      & 1 / 4  & 6 / 7  & 2 / 12  & 9 / 23    & 3 / 11  & 0 / 16 & 4 / 13  & 7 / 12  & 32             & 66             & 0.33              \\
Sage         & 5 / 5  & 6 / 7  & 2 / 15  & 12 / 27   & 14 / 28 & 0 / 20 & 47 / 61 & 9 / 18  & 95    & 86             & {0.52}        \\
davinci-text-001   & 4 / 5  & 4 / 6  & 1 / 4   & 15 / 23   & 7 / 19  & 0 / 0  & 3 / 9   & 0 / 1   & 34             & {33}       & 0.51              \\
davinci-text-002   & 4 / 4  & 6 / 6  & 13 / 28 & 6 / 26    & 8 / 24  & 0 / 4  & 11 / 32 & 9 / 15  & 57             & 82             & 0.41              \\
davinci-text-003   & 4 / 4  & 6 / 7  & 2 / 11  & 6 / 26    & 5 / 17  & 8 / 8  & 3 / 8   & 10 / 18 & 44             & 55             & 0.44              \\
Claude+       & 4 / 5  & 9 / 10 & 1 / 7   & 10 / 15   & 12 / 15 & 0 / 56 & 2 / 14  & 5 / 11  & 43             & 90             & 0.32              \\
Claude-100k & 4 / 4  & 6 / 7  & 6 / 14  & 16 / 20   & 5 / 13  & 0 / 28 & 7 / 12  & 5 / 11  & 49             & 60             & 0.45              \\
GPT-3.5-turbo       & 4 / 6  & 6 / 7  & 6 / 22  & 5 / 8     & 21 / 35 & 1 / 20 & 11 / 27 & 7 / 12  & 61             & 76             & 0.45              \\
GPT-4         & 5 / 5  & 9 / 9  & 8 / 10  & 10 / 10   & 20 / 21 & 9 / 15 & 18 / 27   & 17 / 24 & 95       & 25    & \textbf{0.79}     \\ \bottomrule
\end{tabular}
\begin{flushleft}
    \footnotesize
    For each dataset and LLM, we report the number of LLM-derived causal statements and the number of correct ones. An instance of the value `4 / 5' indicates that an LLM produces 5 causal statements, of which 4 correspond to a directed path in the actual structure.
\end{flushleft}
\label{tab_LLM_perform}
\end{table*}

\begin{table}[!t]
\centering
\caption{Comparative accuracy of LLM-derived causality: As paths v.s. as edges. The better accuracy is highlighted in bold.}
\begin{tabular}{@{}ccccc@{}}
\toprule
\multicolumn{5}{c}{Accuracy of Paths / Accuracy of Edges w.r.t. LLM-derived causality} \\ \midrule
Dataset    & GPT-4                 & GPT-3.5-turbo        & Claude+              & Claude-100k          \\ \midrule
Cancer     & \textbf{1.00} / 0.80  & 0.67 / 0.67          & 0.80 / 0.80          & 1.00 / 1.00          \\
Asia       & \textbf{1.00} / 0.78  & 0.86 / 0.86          & \textbf{0.90} / 0.70 & \textbf{0.86} / 0.57 \\
Child      & \textbf{0.80} / 0.70  & 0.27 / 0.27          & \textbf{0.14} / 0.00 & 0.43 / 0.43          \\
Insurance  & \textbf{1.00} / 0.70  & \textbf{0.63} / 0.38 & \textbf{0.67} / 0.60 & \textbf{0.80} / 0.60 \\
Alarm      & \textbf{0.95} / 0.67  & \textbf{0.60} / 0.51 & \textbf{0.80} / 0.27 & 0.38 / 0.38          \\
Water      & 0.60 / 0.60           & \textbf{0.05} / 0.00 & 0.00 / 0.00          & 0.00 / 0.00          \\
Mildew     & \textbf{0.67} / 0.48  & \textbf{0.41} / 0.22 & 0.14 / 0.14          & \textbf{0.58} / 0.50 \\
Barley     & \textbf{0.71} / 0.46  & \textbf{0.58} / 0.33 & \textbf{0.45} / 0.36 & \textbf{0.45} / 0.36 \\ \bottomrule
\end{tabular}
\label{tab_LLM_structure}
\end{table}

This study utilizes eight datasets from the Bayesian Network repository\footnote{\url{https://www.bnlearn.com/bnrepository/}} to conduct experiments. These datasets span various domains and complexities and are detailed in terms of variables, edges, parameters, directed paths (which represent qualitative causality), and domains of application. The specific characteristics of each dataset are presented in Table \ref{tab_datasets}. For our experiments, we employ observational data as provided by Li \textit{et al.} \cite{li2018bayesian}. This synthetic data comprises 12 groups, each containing two subsets of differing sizes, across all datasets.
In addition, we use two private datasets that were not published, thus not included in the training of LLMs for evaluation. The first dataset, RECTUM, contains clinical data from rectal cancer patients and includes nine variables. The second dataset, HEAD\&NECK, involves data from patients with head and neck tumors, comprising fourteen variables.

For LLM-based causal reasoning, we use a range of models operational as of June 2023, including GPT-3.5-turbo, GPT-4 \cite{achiam2023gpt}, Claude+ \cite{Claude}, Claude-100k \cite{Claude100k}, various GPT text models\footnote{\url{https://platform.openai.com/docs/api-reference}}, ChatGLM \cite{GLM}, Sage \cite{sage}, Chatsonic\footnote{\url{https://writesonic.com/}}, HuggingChat\footnote{\url{https://huggingface.co/chat/}}, Whispr\footnote{\url{https://whispr.chat/}}, Xinhuo\footnote{\url{https://passport.xfyun.cn/}}, and Yiyan\footnote{\url{https://yiyan.baidu.com/welcome}}. For structure learning, we employ the Minimum Message Length (MML) \cite{o2006learning} scoring function, an instance of the MDL scoring criteria, and MCMC search \cite{o2006learning} in the soft approach and the BDeu \cite{heckerman1995learning} scoring function and Tabu search \cite{lee2017metaheuristics} in the hard one.

The experiments are executed on a single core of an AMD Ryzen9 7950x CPU with a clock speed of 4.5GHz.

\subsection{Evaluation Metrics}
We utilize the structural Hamming distance (SHD) \cite{chickering2002learning}, F1 score and structural intervention distance (SID) \cite{peters2015structural,henckel2024adjustment,cohrs2025large} to quantify the quality of recovered causal structure\footnote{The reported results use DAGs rather than completed partially DAGs, which represent Markov Equivalence Classes, because we consider the causal DAG as a knowledge-level outcome in which the orientation of all edges is essential for understanding causal mechanisms.}. 

The SHD represents the discrepancy in edges between the inferred and actual causal DAGs, with a lower SHD indicating superior model performance.
The F1 score is computed as \(\frac{2 \cdot \text{precision} \cdot \text{recall}}{\text{precision} + \text{recall}}\). In the context of structure learning, precision denotes the proportion of correctly identified edges to all edges identified by the approach. Recall is the fraction of accurately identified edges relative to the total number of edges in the actual causal structure.
The SID quantifies the closeness between two DAGs in terms of causal inference statements.

\subsection{Comparative Analysis}

We conduct a comparative analysis between the LLM-driven CSL and the traditional data-based CSL across eight datasets. This comparison leverages the methodologies outlined in Sections \ref{sec_method_anc}, \ref{sec_method_soft_search} (for the soft approach), and Section \ref{sec_method_hard} (for the hard approach). The comparative outcomes for the soft and hard approaches are reported in Tables \ref{tab_cmp_soft}, \ref{tab_cmp_hard}, and \ref{tab:SID_comp}.
GPT-4 is used to derive prior constraints, and the confidence level\footnote{Refer to Section \ref{sec:confidence} for the discussion on confidence level settings.} is set to $0.99999$.
For comparison purpose, we present the relative performance difference, calculated by $(\Delta_{\text{LLM}} - \Delta_{\text{Data}})/\Delta_{\text{Data}}$.
Here, \(\Delta_{\text{LLM}}\) and \(\Delta_{\text{Data}}\) denote the performances of the LLM-driven and data-based CSL approaches, respectively. Instances where the LLM-driven CSL outperforms the data-based approach are highlighted in red or bold.

 \begin{figure*}[!t]
    \centering
    \includegraphics[width=0.95\textwidth]{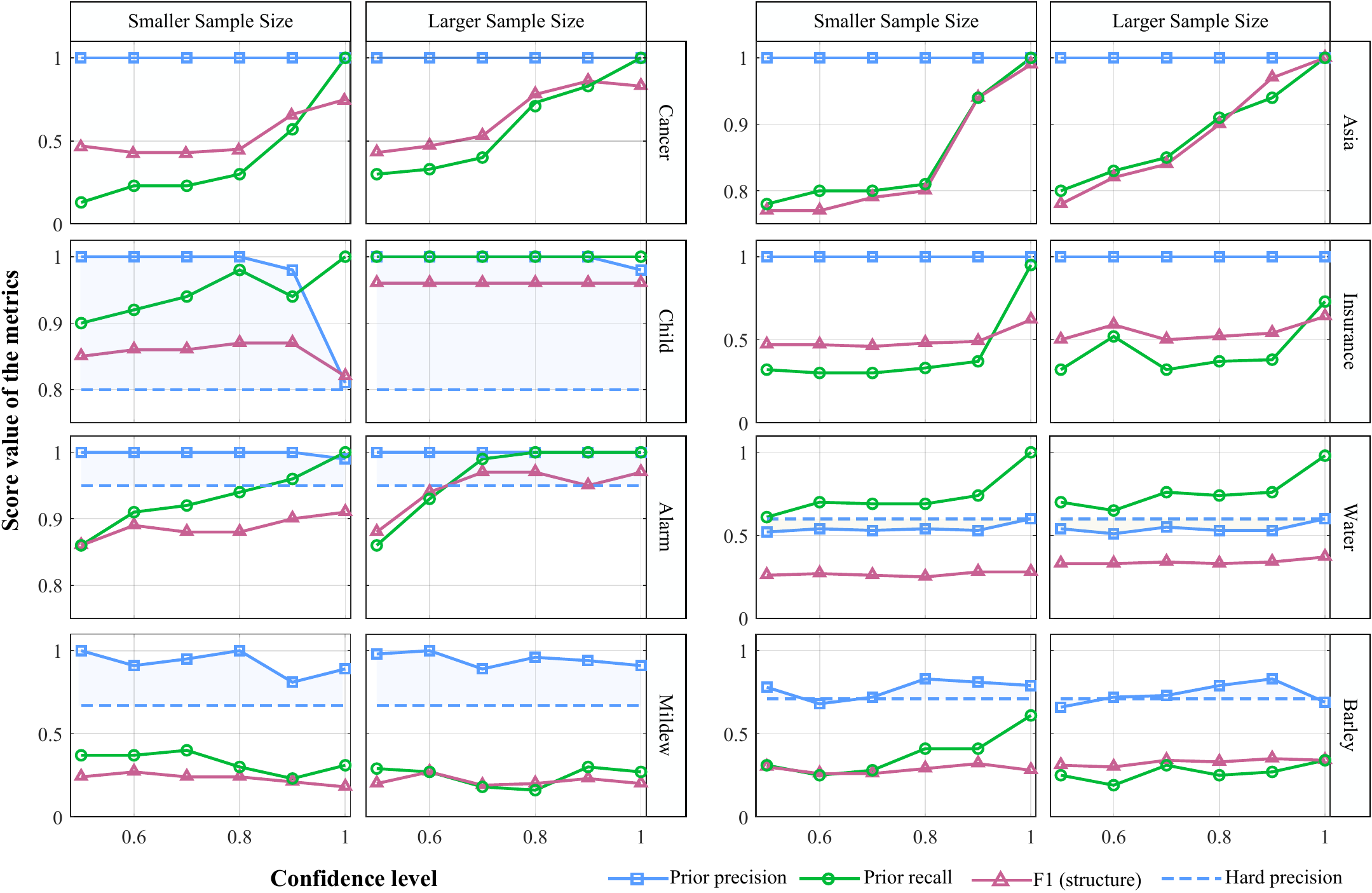}
    \caption{Results of the soft approach with varying confidence levels in terms of the precision and recall of accepted priors, and the F1 score of the recovered causal structure. The dotted blue line represents the accuracy of LLM-derived priors, equivalent to the prior precision in the hard approach. The colored areas illustrate changes in prior precision when utilizing the soft approach compared to the original accuracy of LLM-derived priors.}
    \label{fig:soft_confidence}
\end{figure*}

Observations reveal that the soft approach generally yields more consistent enhancements in CSL performance compared to the hard approach. This is particularly notable as over half of the instances exhibit a decline in performance when applying LLM-derived priors via the hard approach, whereas the soft approach typically enhances the performance of data-based CSL. This discrepancy is attributed to the imperfections in the LLM-derived prior constraints, as detailed in Table \ref{tab_LLM_perform}. The soft approach's error tolerance effectively mitigates the negative impact of erroneous constraints, thereby preserving or enhancing performance despite the presence of inaccuracies in LLM-derived prior constraints.

\subsection{Quality of LLM-derived Prior as Different Structures}

We investigate various LLMs on deriving causal statements with our prompting strategy.
The number of LLM-derived statements, along with the number of ones that correspond to a path\footnote{Here, the path includes both edges and indirect paths.} in the actual causal structure, is reported in  Table \ref{tab_LLM_perform}.
To further investigate the rationale of our use of LLM-derived priors as ancestral constraints, we compare the accuracy of them as paths and edges, with results reported in Table \ref{tab_LLM_structure}.

For the results outlined in Table \ref{tab_LLM_perform}, we observe that GPT-4 excels in deriving accurate causal statements, achieving an overall accuracy rate of 0.79. The performance of other models varies, with accuracy rates ranging from 0.16 to 0.56. In smaller datasets, such as \textit{Cancer} and \textit{Asia}, most LLMs demonstrate high accuracy in deriving causal statements. However, in larger datasets, the performance of most LLMs is inconsistent, excelling in some domains but underperforming in others. Notably, GPT-4 exhibits stable performance across all datasets, indicating its robustness in capturing detailed metadata information across diverse domains.

For the results presented in Table \ref{tab_LLM_structure}, we observe that the accuracy of paths with respect to LLM-derived causality is generally higher than that of edges, particularly for GPT-4. This finding supports our approach of using LLM-derived causal statements as ancestral constraints (existence of paths) rather than directly as edge existence constraints.

\begin{table*}[!h]
\centering
\caption{Representative examples of erroneous prior causality derived from GPT-4.}
\begin{tabular}{l|ll|p{5cm}}
\hline
Dataset                 & Variable   & Description                                                           & Erroneous Causal Statements                                                                                                                                                                                                       \\ \hline
\multirow{4}{*}{Child}  & HypoxiaInO2 & Severity of hypoxia                                                   & \multirow{4}{3.5cm}{(HypoxiaInO2,Grunting); (LungFlow,HypDistrib)}                                                                                                                                                                 \\
                        & LungFlow    & Blood flow in the lungs                                               &                                                                                                                                                                                                                                \\
                        & Grunting    & Presence of grunting                                                  &                                                                                                                                                                                                                                \\
                        & HypDistrib  & Distribution of blood flow in the body                                &                                                                                                                                                                                                                                \\ \hline
\multirow{2}{*}{Alarm}  & HRSAT       & Heart Rate Saturation                                                 & \multirow{2}{6.5cm}{(HRSAT,HR)}                                                                                                                                                                                                    \\
                        & HR          & Heart Rate                                                            &                                                                                                                                                                                                                                \\ \hline
\multirow{4}{*}{Barley} & forfrugt    & Preceding crop                                                        & \multirow{4}{3cm}{(forfrugt,nmin); (ngodnt,nopt)}                                                                                                                                                                                 \\
                        & ngodnt      & Nitrogen content in the grain at optimal nitrogen treatment           &                                                                                                                                                                                                                                \\
                        & nmin        & Mineral nitrogen content in the soil                                  &                                                                                                                                                                                                                                \\
                        & nopt        & Optimal nitrogen content                                              &                                                                                                                                                                                                                                \\ \hline
\multirow{4}{*}{Water}  & CKNI\_H\_M  & Concentration of Kjeldahl nitrogen inorganic  ions at hour H minute M & \multirow{4}{6.5cm}{(CKNI\_12\_00,C\_NI\_12\_15); (CKND\_12\_00,CNOD\_12\_15); (CKNI\_12\_15,C\_NI\_12\_30); (CKND\_12\_15,CNOD\_12\_30)}                                                                                         \\
                        & C\_NI\_H\_M & Number of nitrogen inorganic ions at hour H minute M                  &                                                                                                                                                                                                                                \\
                        & CKND\_H\_M  & Concentration of Kjeldahl nitrogen degradable ions at hour H minute M &                                                                                                                                                                                                                                \\
                        & CNOD\_H\_M  & Concentration of Kjeldahl nitrogen degradable ions at hour H minute M &                                                                                                                                                                                                                                \\ \hline
\multirow{4}{*}{Mildew} & mikro\_k    & Microclimate conditions at stage k                                    & \multirow{4}{7cm}{(mikro\_1,meldug\_1); (middel\_1,meldug\_1); (nedboer\_1,meldug\_1); (mikro\_2,meldug\_2); (middel\_2,meldug\_2); (nedboer\_2,meldug\_3)} \\
                        & meldug\_k   & Degree of mildew infestation at stage k                               &                                                                                                                                                                                                                                \\
                        & middel\_k   & Average amount of fungicide applied stage k                           &                                                                                                                                                                                                                                \\
                        & nedboer\_k  & Amount of precipitation stage k                                       &                                                                                                                                                                                                                                \\ \hline
\end{tabular}
\label{erroneousCausal}
\end{table*}

\subsection{Error Tolerance of the Soft Approach}

In this experiment, we investigate the error-tolerant capability of the soft approach and the impact of varying confidence levels on it. For the LLM-derived prior constraint set, we report the precision and recall rates of the constraints accepted by the soft approach under different confidence levels, ranging from $0.5$ to $0.99999$. We also compare this precision with the original precision of the constraints (referred to as hard precision). Additionally, we evaluate the changes in the output structure quality using F1 scores to assess the impact of the quality of accepted priors on the quality of recovered causality. These results are depicted in Figure \ref{fig:soft_confidence}.

We observe that
1) In most cases, the precision of accepted priors is higher than their original accuracy (hard precision), demonstrating the effectiveness of filtering prior errors.
2) In some cases, there is an increasing trend in structural quality as the confidence level increases. This is evidenced by consistently high precision of accepted priors and increasing recall rates, as seen in the \textit{Alarm} and \textit{Asia} datasets.
3) Conversely, in some cases such as the \textit{Child} and \textit{Barley} datasets, the precision of accepted priors decreases as the confidence level increases from 0.9 to 0.99999, where some erroneous priors are forced to be accepted, leading to decreased structural quality.

These findings underscore the soft approach's error tolerance, effectively filtering out erroneous priors and ensuring the accuracy of the accepted ones. Additionally, setting the confidence level properly is crucial for optimal performance. In some instances, a confidence level of 0.9 outperforms the previously used 0.99999 setting. Confidence levels can be adjusted based on practical scenarios, with lower levels suitable for high-quality observational data and higher levels needed for lower-quality data.

\begin{figure*}[!h]
    \centering
    \includegraphics[width=0.98 \textwidth]{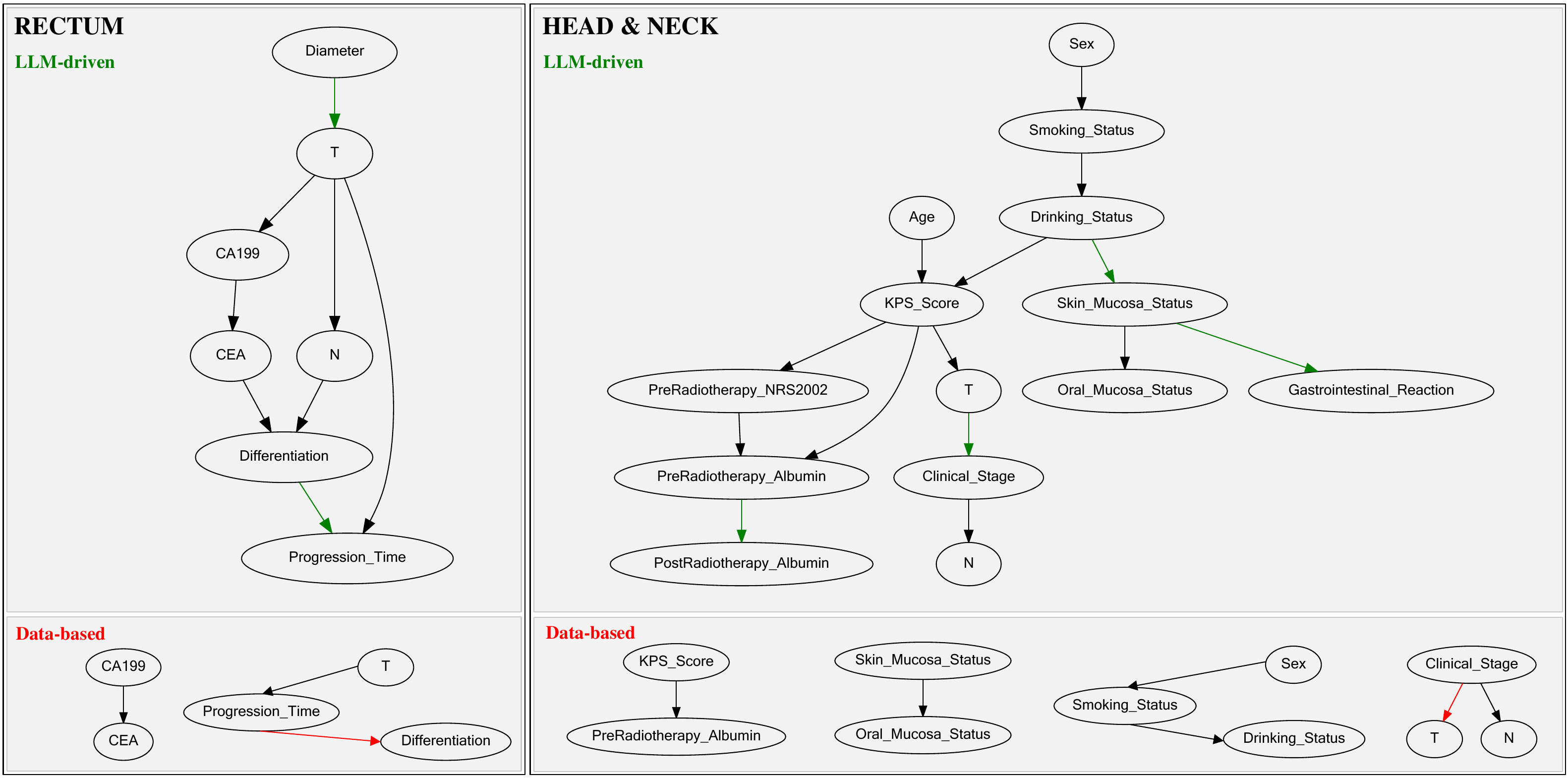}
    \caption{Visualization of LLM-driven and data-based structure learning outcomes on private data. The recovered missing causality is highlighted by \textcolor{green2}{GREEN} arrows, while the removed erroneous causality is highlighted by \textcolor{red2}{RED} arrows.}
    \label{fig:real_data}
\end{figure*}

\subsection{Errors in GPT4-derived Causal Statements}

In this examination, we detail the errors made by GPT-4 in inferring causal statements, as highlighted in Table \ref{erroneousCausal}. The predominant categories of these mistakes include: 1) \textbf{Unclear Causal Mechanism.} A significant source of error arises from GPT-4's inherent difficulty in interpreting unclear mechanisms. For example, in the \textit{Barley} dataset, there appears to be a qualitative causality from nitrogen content in grain ($\textit{ngodnt}$) to optimal nitrogen content ($\textit{nopt}$). However, this perceived causality is actually driven by other interaction mechanisms rather than a directed causal path.
2) \textbf{Insufficient Metadata.} Errors also arise when GPT-4 adopts a too-generalized view, misaligning with the dataset’s specific context due to limited contextual or metadata information. In the \textit{Child} dataset, the model inaccurately posits that low oxygen levels ($\textit{HypoxiaInO2}$) lead to $\textit{Grunting}$, ignoring the dataset’s specific focus on infant hypoxia conditions and the nuanced relationships between symptoms and underlying health issues.

These errors underscore the challenges faced by GPT-4 in accurately extracting causal knowledge that aligns with the actual structure, especially for unclear mechanisms or insufficient metadata. To enhance the performance of LLM-driven causal discovery, it is crucial to use LLM-derived priors more carefully in structure learning and to provide more sufficient and clear metadata.

\subsection{Results on Private Real-World Data}
This experiment compares LLM-driven causal discovery with data-based methods on private real-world datasets that were not part of the LLM's training data, thereby evaluating the generalizability of our approach. We present the visualized structures obtained using CaMML as the backbone algorithm, highlighting the distinct edges between our LLM-driven and data-based causal discovery methods in Fig. \ref{fig:real_data}.

The LLM-driven causal discovery results show a more connected and meaningful causal structure compared to the data-based approach. Notably, even without explicit edge constraints, LLM-driven structure learning effectively recovers missing edges representing causal mechanisms that cannot be identified by data alone, while also removing erroneous edges without being explicitly instructed by priors. 
These results demonstrate the strong generalizability of our LLM-driven approach and its effectiveness in handling real-world data.

\section{Discussions}
\label{sec:conclusions}

\subsection{Conclusions}

This paper introduces a novel approach to causal discovery by integrating  LLMs to provide prior constraints. We design a three-fold error-tolerant mechanism to address the inaccuracies of LLM-derived priors in revealing the actual causal structure.
Specifically, we propose 1) an accuracy-oriented prompting strategy that adapts LLM-based causal reasoning to the causal discovery context, 2) a knowledge-to-structure transition that aligns LLM-derived causality with its qualitative nature, and 3) a soft approach for integrating LLM-derived priors in structure learning, introducing tolerance for remaining prior errors that escape the previous two steps.
The experimental results demonstrate the promise of employing LLMs as innovative tools to advance causal discovery. We expect to inspire further exploration of combining LLMs with causal research.

\subsection{Limitations and Future Directions}

Despite promising results in many cases, limitations remain in the current strategy of using LLMs to augment data-based causal discovery. Even when employing looser structural constraints, such as modeling LLM-derived priors through the existence of paths rather than edges, inaccurate priors can still negatively impact structure learning. This issue is particularly pronounced in temporal contexts, as seen in the \textit{Water} and \textit{Mildew} datasets, where the performance of MINOBSx is notably affected.
Additionally, our focus has primarily been on combinatorial structure learning for causal discovery. The integration of LLM-derived priors within continuous structure learning frameworks \cite{zheng2018dags} remains largely unexplored, particularly for ancestral constraints, for which no integration methods currently exist.

Therefore, future research can aim to improve LLM accuracy in identifying causal relationships, especially in distinguishing between direct and indirect influences. Enhancing methods for extracting high-quality causal statements from LLMs could also boost interpretability. Moreover, exploring alternative forms of prior constraints and developing algorithms that more effectively filter reliable information from LLM-generated data could significantly increase the robustness and trustworthiness of LLM-augmented causal discovery.

\subsection{Emerging LLM-driven Causal Discovery Studies}

Since this study was first conducted, a series of studies have proposed various strategies to leverage LLM-based causal reasoning for aiding causal discovery.

Several studies have introduced iterative approaches to integrate LLM-derived causal reasoning into structure learning. Abdulaal et al. \cite{abdulaal2023causal} proposed a causal agent by prompting pairwise causal relationships among variables as edge constraints to inform structure learning. Ban et al. \cite{ban2023causal} restricted pairwise causal reasoning within the bounds of data-implied causal structures, showing improved accuracy in priors. Takayama et al. \cite{takayama2024integrating} further refined the integration between LLM-based causal reasoning and data-driven causal discovery, proposing advanced prompting and knowledge integration strategies.

Vashishtha et al. \cite{vashishtha2023causal,vashishtha2024causal} focused on extracting causal orders, a crucial information for causal discovery, by designing effective LLM prompting strategies to derive variable orderings, thereby reducing the complexity of causal discovery. The authors also introduced an innovative axiomatic training scheme to enhance LLMs' causal reasoning abilities \cite{vashishtha2024teaching}. Additionally, Ban et al. \cite{ban2024differentiable} explored how to incorporate partial orders into differentiable structure learning, unlocking the potential of LLM-driven differentiable causal discovery.

In the context of dynamic causal discovery, some studies have utilized LLM-based causal reasoning to focus on temporal causal relationships, employing specialized prompting schemes \cite{li2024realtcd,li2024llm,lee2024incorporating}. Other studies have developed end-to-end LLM-driven causal discovery tools that integrate data-based structure learning, using code interpreters to enable LLMs to automatically invoke causal learning packages \cite{jiangllm4causal,khatibi2024alcm}.

Additional explorations include using LLMs to discover hidden variables in datasets \cite{liu2024discovery} and building foundational models for causal inference \cite{zhangtowards}. For a comprehensive overview, related surveys are available that cover various aspects of using LLMs for causality \cite{liu2024large,sun2023survey,ma2024causal,zhou2023emerging}.

\bibliographystyle{IEEEtran}
\footnotesize\bibliography{ref}

\begin{thebibliography}{10}
\providecommand{\url}[1]{#1}
\csname url@samestyle\endcsname
\providecommand{\newblock}{\relax}
\providecommand{\bibinfo}[2]{#2}
\providecommand{\BIBentrySTDinterwordspacing}{\spaceskip=0pt\relax}
\providecommand{\BIBentryALTinterwordstretchfactor}{4}
\providecommand{\BIBentryALTinterwordspacing}{\spaceskip=\fontdimen2\font plus
\BIBentryALTinterwordstretchfactor\fontdimen3\font minus \fontdimen4\font\relax}
\providecommand{\BIBforeignlanguage}[2]{{%
\expandafter\ifx\csname l@#1\endcsname\relax
\typeout{** WARNING: IEEEtran.bst: No hyphenation pattern has been}%
\typeout{** loaded for the language `#1'. Using the pattern for}%
\typeout{** the default language instead.}%
\else
\language=\csname l@#1\endcsname
\fi
#2}}
\providecommand{\BIBdecl}{\relax}
\BIBdecl

\bibitem{sachs2005causal}
K.~Sachs, O.~Perez, D.~Pe'er, D.~A. Lauffenburger, and G.~P. Nolan, ``Causal protein-signaling networks derived from multiparameter single-cell data,'' \emph{Science}, vol. 308, no. 5721, pp. 523--529, 2005.

\bibitem{kyono2021exploiting}
T.~Kyono and M.~Van~der Schaar, ``Exploiting causal structure for robust model selection in unsupervised domain adaptation,'' \emph{IEEE Transactions on Artificial Intelligence}, vol.~2, no.~6, pp. 494--507, 2021.

\bibitem{mascaro2023modeling}
S.~Mascaro, Y.~Wu, O.~Woodberry, E.~P. Nyberg, R.~Pearson, J.~A. Ramsay, A.~O. Mace, D.~A. Foley, T.~L. Snelling, A.~E. Nicholson \emph{et~al.}, ``Modeling {COVID-19} disease processes by remote elicitation of causal bayesian networks from medical experts,'' \emph{BMC Medical Research Methodology}, vol.~23, no.~1, p.~76, 2023.

\bibitem{yang2022online}
J.~Yang, L.~Jiang, A.~Shen, and A.~Wang, ``Online streaming features causal discovery algorithm based on partial rank correlation,'' \emph{IEEE Transactions on Artificial Intelligence}, vol.~4, no.~1, pp. 197--208, 2022.

\bibitem{chen2022directed}
Z.~Chen and Z.~Ge, ``Directed acyclic graphs with tears,'' \emph{IEEE Transactions on Artificial Intelligence}, vol.~4, no.~4, pp. 972--983, 2022.

\bibitem{moe2021increased}
S.~J. Moe, J.~F. Carriger, and M.~Glendell, ``Increased use of {Bayesian} network models has improved environmental risk assessments,'' \emph{Integrated Environmental Assessment and Management}, vol.~17, no.~1, pp. 53--61, 2021.

\bibitem{scutari2019learns}
M.~Scutari, C.~E. Graafland, and J.~M. Guti{\'e}rrez, ``Who learns better {Bayesian} network structures: Accuracy and speed of structure learning algorithms,'' \emph{International Journal of Approximate Reasoning}, vol. 115, pp. 235--253, 2019.

\bibitem{xiang2023bootstrap}
G.~Xiang, H.~Wang, K.~Yu, X.~Guo, F.~Cao, and Y.~Song, ``Bootstrap-based layerwise refining for causal structure learning,'' \emph{IEEE Transactions on Artificial Intelligence}, vol.~5, no.~6, pp. 2708--2722, 2024.

\bibitem{vowels2022d}
M.~J. Vowels, N.~C. Camgoz, and R.~Bowden, ``D’ya like {DAGs}? {A} survey on structure learning and causal discovery,'' \emph{ACM Computing Surveys}, vol.~55, no.~4, pp. 1--36, 2022.

\bibitem{heinze2018causal}
C.~Heinze-Deml, M.~H. Maathuis, and N.~Meinshausen, ``Causal structure learning,'' \emph{Annual Review of Statistics and Its Application}, vol.~5, pp. 371--391, 2018.

\bibitem{chickering1996learning}
D.~M. Chickering, ``Learning bayesian networks is {NP-complete},'' \emph{Learning from data: Artificial intelligence and statistics V}, pp. 121--130, 1996.

\bibitem{heckerman1995learning2}
D.~Heckerman, D.~Geiger, and D.~M. Chickering, ``Learning bayesian networks: The combination of knowledge and statistical data,'' \emph{Machine Learning}, vol.~20, pp. 197--243, 1995.

\bibitem{de2011efficient}
C.~P. De~Campos and Q.~Ji, ``Efficient structure learning of {Bayesian} networks using constraints,'' \emph{The Journal of Machine Learning Research}, vol.~12, pp. 663--689, 2011.

\bibitem{amirkhani2016exploiting}
H.~Amirkhani, M.~Rahmati, P.~J. Lucas, and A.~Hommersom, ``Exploiting experts{'} knowledge for structure learning of {Bayesian} networks,'' \emph{IEEE Transactions on Pattern Analysis and Machine Intelligence}, vol.~39, no.~11, pp. 2154--2170, 2016.

\bibitem{constantinou2023impact}
A.~C. Constantinou, Z.~Guo, and N.~K. Kitson, ``The impact of prior knowledge on causal structure learning,'' \emph{Knowledge and Information Systems}, pp. 1--50, 2023.

\bibitem{maydeu2020estimating}
A.~Maydeu-Olivares, D.~Shi, and A.~J. Fairchild, ``Estimating causal effects in linear regression models with observational data: The instrumental variables regression model.'' \emph{Psychological Methods}, vol.~25, no.~2, p. 243, 2020.

\bibitem{zhang2023causality}
C.~Zhang, D.~Janzing, M.~van~der Schaar, F.~Locatello, and P.~Spirtes, ``Causality in the time of {LLMs}: Round table discussion results of clear 2023,'' \emph{Proceedings of Machine Learning Research vol TBD}, vol.~1, p.~7, 2023.

\bibitem{kiciman2023causal}
E.~K{\i}c{\i}man, R.~Ness, A.~Sharma, and C.~Tan, ``Causal reasoning and large language models: Opening a new frontier for causality,'' \emph{arXiv preprint arXiv:2305.00050}, 2023.

\bibitem{nori2023capabilities}
H.~Nori, N.~King, S.~M. McKinney, D.~Carignan, and E.~Horvitz, ``Capabilities of {GPT-4} on medical challenge problems,'' \emph{arXiv preprint arXiv:2303.13375}, 2023.

\bibitem{long2023can}
S.~Long, T.~Schuster, A.~Pich{\'e}, S.~Research \emph{et~al.}, ``Can large language models build causal graphs?'' \emph{arXiv preprint arXiv:2303.05279}, 2023.

\bibitem{zevcevic2023causal}
M.~Ze{\v{c}}evi{\'c}, M.~Willig, D.~S. Dhami, and K.~Kersting, ``Causal parrots: Large language models may talk causality but are not causal,'' \emph{arXiv preprint arXiv:2308.13067}, 2023.

\bibitem{lu2023emergent}
S.~Lu, I.~Bigoulaeva, R.~Sachdeva, H.~T. Madabushi, and I.~Gurevych, ``Are emergent abilities in large language models just in-context learning?'' \emph{arXiv preprint arXiv:2309.01809}, 2023.

\bibitem{tu2023causal}
R.~Tu, C.~Ma, and C.~Zhang, ``Causal-discovery performance of {chatGPT} in the context of neuropathic pain diagnosis,'' \emph{arXiv preprint arXiv:2301.13819}, 2023.

\bibitem{wei2022chain}
J.~Wei, X.~Wang, D.~Schuurmans, M.~Bosma, F.~Xia, E.~Chi, Q.~V. Le, D.~Zhou \emph{et~al.}, ``Chain-of-thought prompting elicits reasoning in large language models,'' in \emph{Advances in Neural Information Processing Systems}, vol.~35, 2022, pp. 24\,824--24\,837.

\bibitem{zhang2023understanding}
C.~Zhang, S.~Bauer, P.~Bennett, J.~Gao, W.~Gong, A.~Hilmkil, J.~Jennings, C.~Ma, T.~Minka, N.~Pawlowski \emph{et~al.}, ``Understanding causality with large language models: Feasibility and opportunities,'' \emph{arXiv preprint arXiv:2304.05524}, 2023.

\bibitem{jin2023can}
Z.~Jin, J.~Liu, L.~Zhiheng, S.~Poff, M.~Sachan, R.~Mihalcea, M.~T. Diab, and B.~Sch{\"o}lkopf, ``Can large language models infer causation from correlation?'' in \emph{The Twelfth International Conference on Learning Representations}, 2023.

\bibitem{willig2022can}
M.~Willig, M.~Ze{\v{c}}evi{\'c}, D.~S. Dhami, and K.~Kersting, ``Can foundation models talk causality?'' \emph{arXiv preprint arXiv:2206.10591}, 2022.

\bibitem{hoyer2008nonlinear}
P.~Hoyer, D.~Janzing, J.~M. Mooij, J.~Peters, and B.~Sch{\"o}lkopf, ``Nonlinear causal discovery with additive noise models,'' \emph{Advances in Neural Information Processing Systems}, vol.~21, 2008.

\bibitem{frohberg2021crass}
J.~Frohberg and F.~Binder, ``Crass: A novel data set and benchmark to test counterfactual reasoning of large language models,'' \emph{arXiv preprint arXiv:2112.11941}, 2021.

\bibitem{choi2022lmpriors}
K.~Choi, C.~Cundy, S.~Srivastava, and S.~Ermon, ``{LMPriors}: Pre-trained language models as task-specific priors,'' in \emph{Proceedings of the NeurIPS 2022 Foundation Models for Decision Making Workshop}, 2022.

\bibitem{pearl2009causality}
J.~Pearl, \emph{Causality}.\hskip 1em plus 0.5em minus 0.4em\relax Cambridge university press, 2009.

\bibitem{zheng2018dags}
X.~Zheng, B.~Aragam, P.~K. Ravikumar, and E.~P. Xing, ``{DAGs} with no tears: Continuous optimization for structure learning,'' \emph{Advances in Neural Information Processing Systems}, vol.~31, 2018.

\bibitem{yang2024additive}
J.~Yang, T.~Lu, and Y.~Zhu, ``Additive noise model structure learning based on spatial coordinates,'' \emph{IEEE Transactions on Artificial Intelligence}, 2024.

\bibitem{de2009structure}
C.~P. De~Campos, Z.~Zeng, and Q.~Ji, ``Structure learning of {Bayesian} networks using constraints,'' in \emph{Proceedings of the 26th Annual International Conference on Machine Learning}, 2009, pp. 113--120.

\bibitem{scutari2017bayesian}
M.~Scutari, ``{Bayesian} network constraint-based structure learning algorithms: Parallel and optimized implementations in the bnlearn {R} package,'' \emph{Journal of Statistical Software}, vol.~77, pp. 1--20, 2017.

\bibitem{bouchaala2010improving}
L.~Bouchaala, A.~Masmoudi, F.~Gargouri, and A.~Rebai, ``Improving algorithms for structure learning in {Bayesian} networks using a new implicit score,'' \emph{Expert Systems with Applications}, vol.~37, no.~7, pp. 5470--5475, 2010.

\bibitem{gasse2012experimental}
M.~Gasse, A.~Aussem, and H.~Elghazel, ``An experimental comparison of hybrid algorithms for {Bayesian} network structure learning,'' in \emph{Joint European Conference on Machine Learning and Knowledge Discovery in Databases}.\hskip 1em plus 0.5em minus 0.4em\relax Springer, 2012, pp. 58--73.

\bibitem{li2018bayesian}
A.~Li and P.~Beek, ``{Bayesian} network structure learning with side constraints,'' in \emph{International Conference on Probabilistic Graphical Models}.\hskip 1em plus 0.5em minus 0.4em\relax PMLR, 2018, pp. 225--236.

\bibitem{schwarz1978estimating}
G.~Schwarz, ``Estimating the dimension of a model,'' \emph{The Annals of Statistics}, vol.~6, pp. 461--464, 1978.

\bibitem{rissanen1978modeling}
J.~Rissanen, ``Modeling by shortest data description,'' \emph{Automatica}, vol.~14, no.~5, pp. 465--471, 1978.

\bibitem{buntine1991theory}
W.~Buntine, ``Theory refinement on {Bayesian} networks,'' in \emph{Uncertainty Proceedings}.\hskip 1em plus 0.5em minus 0.4em\relax Elsevier, 1991, pp. 52--60.

\bibitem{heckerman1995learning}
D.~Heckerman and D.~Geiger, ``Learning {Bayesian} networks: A unification for discrete and gaussian domains,'' in \emph{Proceedings of the 11th Conference on Uncertainty in Artificial Intelligence}, 1995, pp. 274--284.

\bibitem{chickering2002optimal}
D.~M. Chickering, ``Optimal structure identification with greedy search,'' \emph{Journal of Machine Learning Research}, vol.~3, no. Nov, pp. 507--554, 2002.

\bibitem{yuan2011learning}
C.~Yuan, B.~Malone, and X.~Wu, ``Learning optimal {Bayesian} networks using {A*} search,'' in \emph{Proceedings of the 21nd International Joint Conference on Artificial Intelligence}, 2011, pp. 2186--2191.

\bibitem{o2006learning}
R.~T. O’Donnell, L.~Allison, and K.~B. Korb, ``Learning hybrid {Bayesian} networks by {MML},'' in \emph{AI 2006: Advances in Artificial Intelligence: 19th Australian Joint Conference on Artificial Intelligence}.\hskip 1em plus 0.5em minus 0.4em\relax Springer, 2006, pp. 192--203.

\bibitem{zhao2023enhancing}
X.~Zhao, M.~Li, W.~Lu, C.~Weber, J.~H. Lee, K.~Chu, and S.~Wermter, ``Enhancing zero-shot chain-of-thought reasoning in large language models through logic,'' \emph{arXiv preprint arXiv:2309.13339}, 2023.

\bibitem{zhou2022least}
D.~Zhou, N.~Sch{\"a}rli, L.~Hou, J.~Wei, N.~Scales, X.~Wang, D.~Schuurmans, C.~Cui, O.~Bousquet, Q.~V. Le \emph{et~al.}, ``Least-to-most prompting enables complex reasoning in large language models,'' in \emph{The Eleventh International Conference on Learning Representations}, 2022.

\bibitem{lee2017metaheuristics}
C.~Lee and P.~van Beek, ``Metaheuristics for score-and-search {Bayesian} network structure learning,'' in \emph{Advances in Artificial Intelligence: 30th Canadian Conference on Artificial Intelligence}.\hskip 1em plus 0.5em minus 0.4em\relax Springer, 2017, pp. 129--141.

\bibitem{achiam2023gpt}
OpenAI, ``{GPT-4} technical report,'' \emph{OpenAI}, 2023.

\bibitem{Claude}
Y.-T. Lin and Y.-N. Chen, ``{LLM-Eval}: {U}nified multi-dimensional automatic evaluation for open-domain conversations with large language models,'' \emph{arXiv preprint arXiv:2305.13711}, 2023.

\bibitem{Claude100k}
Z.~Zheng, X.~Ren, F.~Xue, Y.~Luo, X.~Jiang, and Y.~You, ``Response length perception and sequence scheduling: {A}n {LLM}-{E}mpowered {LLM} inference pipeline,'' \emph{arXiv preprint arXiv:2305.13144}, 2023.

\bibitem{GLM}
B.~Yang, A.~Raza, Y.~Zou, and T.~Zhang, ``Customizing general-purpose foundation models for medical report generation,'' \emph{arXiv preprint arXiv:2306.05642}, 2023.

\bibitem{sage}
B.~Y. Lin, Y.~Fu, K.~Yang, P.~Ammanabrolu, F.~Brahman, S.~Huang, C.~Bhagavatula, Y.~Choi, and X.~Ren, ``{SwiftSage}: {A} generative agent with fast and slow thinking for complex interactive tasks,'' \emph{arXiv preprint arXiv:2305.17390}, 2023.

\bibitem{chickering2002learning}
D.~M. Chickering, ``Learning equivalence classes of bayesian-network structures,'' \emph{The Journal of Machine Learning Research}, vol.~2, pp. 445--498, 2002.

\bibitem{peters2015structural}
J.~Peters and P.~B{\"u}hlmann, ``Structural intervention distance for evaluating causal graphs,'' \emph{Neural Computation}, vol.~27, no.~3, pp. 771--799, 2015.

\bibitem{henckel2024adjustment}
L.~Henckel, T.~W{\"u}rtzen, and S.~Weichwald, ``Adjustment identification distance: A gadjid for causal structure learning,'' \emph{arXiv preprint arXiv:2402.08616}, 2024.

\bibitem{cohrs2025large}
K.-H. Cohrs, E.~Diaz, V.~Sitokonstantinou, G.~Varando, and G.~Camps-Valls, ``Large language models for causal hypothesis generation in science,'' \emph{Machine Learning: Science and Technology}, vol.~6, no.~1, p. 013001, 2025.

\bibitem{abdulaal2023causal}
A.~Abdulaal, N.~Montana-Brown, T.~He, A.~Ijishakin, I.~Drobnjak, D.~C. Castro, D.~C. Alexander \emph{et~al.}, ``Causal modelling agents: Causal graph discovery through synergising metadata- and data-driven reasoning,'' in \emph{Proceedings of the 12th International Conference on Learning Representations}, 2023.

\bibitem{ban2023causal}
T.~Ban, L.~Chen, D.~Lyu, X.~Wang, and H.~Chen, ``Causal structure learning supervised by large language model,'' \emph{arXiv preprint arXiv:2311.11689}, 2023.

\bibitem{takayama2024integrating}
M.~Takayama, T.~Okuda, T.~Pham, T.~Ikenoue, S.~Fukuma, S.~Shimizu, and A.~Sannai, ``Integrating large language models in causal discovery: A statistical causal approach,'' \emph{arXiv preprint arXiv:2402.01454}, 2024.

\bibitem{vashishtha2023causal}
A.~Vashishtha, A.~G. Reddy, A.~Kumar, S.~Bachu, V.~N. Balasubramanian, and A.~Sharma, ``Causal inference using {LLM}-guided discovery,'' \emph{arXiv preprint arXiv:2310.15117}, 2023.

\bibitem{vashishtha2024causal}
------, ``Causal order: The key to leveraging imperfect experts in causal inference,'' in \emph{Causality and Large Models@ NeurIPS 2024}, 2024.

\bibitem{vashishtha2024teaching}
A.~Vashishtha, A.~Kumar, A.~Pandey, A.~G. Reddy, V.~N. Balasubramanian, and A.~Sharma, ``Teaching transformers causal reasoning through axiomatic training,'' in \emph{Causality and Large Models@ NeurIPS 2024}, 2024.

\bibitem{ban2024differentiable}
\BIBentryALTinterwordspacing
T.~Ban, L.~Chen, X.~Wang, X.~Wang, D.~Lyu, and H.~Chen, ``Differentiable structure learning with partial orders,'' in \emph{The Thirty-eighth Annual Conference on Neural Information Processing Systems}, 2024. [Online]. Available: \url{https://openreview.net/forum?id=B2cTLakrhV}
\BIBentrySTDinterwordspacing

\bibitem{li2024realtcd}
P.~Li, X.~Wang, Z.~Zhang, Y.~Meng, F.~Shen, Y.~Li, J.~Wang, Y.~Li, and W.~Zhu, ``{RealTCD}: Temporal causal discovery from interventional data with large language model,'' in \emph{Proceedings of the 33rd ACM International Conference on Information and Knowledge Management}, 2024, pp. 4669--4677.

\bibitem{li2024llm}
------, ``{LLM}-enhanced causal discovery in temporal domain from interventional data,'' \emph{arXiv preprint arXiv:2404.14786}, 2024.

\bibitem{lee2024incorporating}
C.~Lee, J.~Kim, Y.~Jeong, Y.~Y. Seok, J.~Lyu, J.-H. Kim, S.~Lee, S.~Han, H.~Choe, S.~Park \emph{et~al.}, ``On incorporating prior knowledge extracted from pre-trained language models into causal discovery,'' in \emph{Causality and Large Models@ NeurIPS 2024}, 2024.

\bibitem{jiangllm4causal}
H.~Jiang, L.~Ge, Y.~Gao, J.~Wang, and R.~Song, ``Llm4causal: Democratized causal tools for everyone via large language model,'' in \emph{First Conference on Language Modeling}, 2024.

\bibitem{khatibi2024alcm}
E.~Khatibi, M.~Abbasian, Z.~Yang, I.~Azimi, and A.~M. Rahmani, ``{ALCM}: Autonomous llm-augmented causal discovery framework,'' \emph{arXiv preprint arXiv:2405.01744}, 2024.

\bibitem{liu2024discovery}
\BIBentryALTinterwordspacing
C.~Liu, Y.~Chen, T.~Liu, M.~Gong, J.~Cheng, B.~Han, and K.~Zhang, ``Discovery of the hidden world with large language models,'' in \emph{The Thirty-eighth Annual Conference on Neural Information Processing Systems}, 2024. [Online]. Available: \url{https://openreview.net/forum?id=w50ICQC6QJ}
\BIBentrySTDinterwordspacing

\bibitem{zhangtowards}
J.~Zhang, J.~Jennings, A.~Hilmkil, N.~Pawlowski, C.~Zhang, and C.~Ma, ``Towards causal foundation model: on duality between optimal balancing and attention,'' in \emph{Forty-first International Conference on Machine Learning}, 2024.

\bibitem{liu2024large}
X.~Liu, P.~Xu, J.~Wu, J.~Yuan, Y.~Yang, Y.~Zhou, F.~Liu, T.~Guan, H.~Wang, T.~Yu \emph{et~al.}, ``Large language models and causal inference in collaboration: A comprehensive survey,'' \emph{arXiv preprint arXiv:2403.09606}, 2024.

\bibitem{sun2023survey}
J.~Sun, C.~Zheng, E.~Xie, Z.~Liu, R.~Chu, J.~Qiu, J.~Xu, M.~Ding, H.~Li, M.~Geng \emph{et~al.}, ``A survey of reasoning with foundation models,'' \emph{arXiv preprint arXiv:2312.11562}, 2023.

\bibitem{ma2024causal}
J.~Ma, ``Causal inference with large language model: A survey,'' \emph{arXiv preprint arXiv:2409.09822}, 2024.

\bibitem{zhou2023emerging}
G.~Zhou, S.~Xie, G.~Hao, S.~Chen, B.~Huang, X.~Xu, C.~Wang, L.~Zhu, L.~Yao, and K.~Zhang, ``Emerging synergies in causality and deep generative models: A survey,'' \emph{arXiv preprint arXiv:2301.12351}, 2023.

\end{thebibliography}

\begin{IEEEbiography}[{\includegraphics[width=1in,height=1.25in,clip,keepaspectratio]{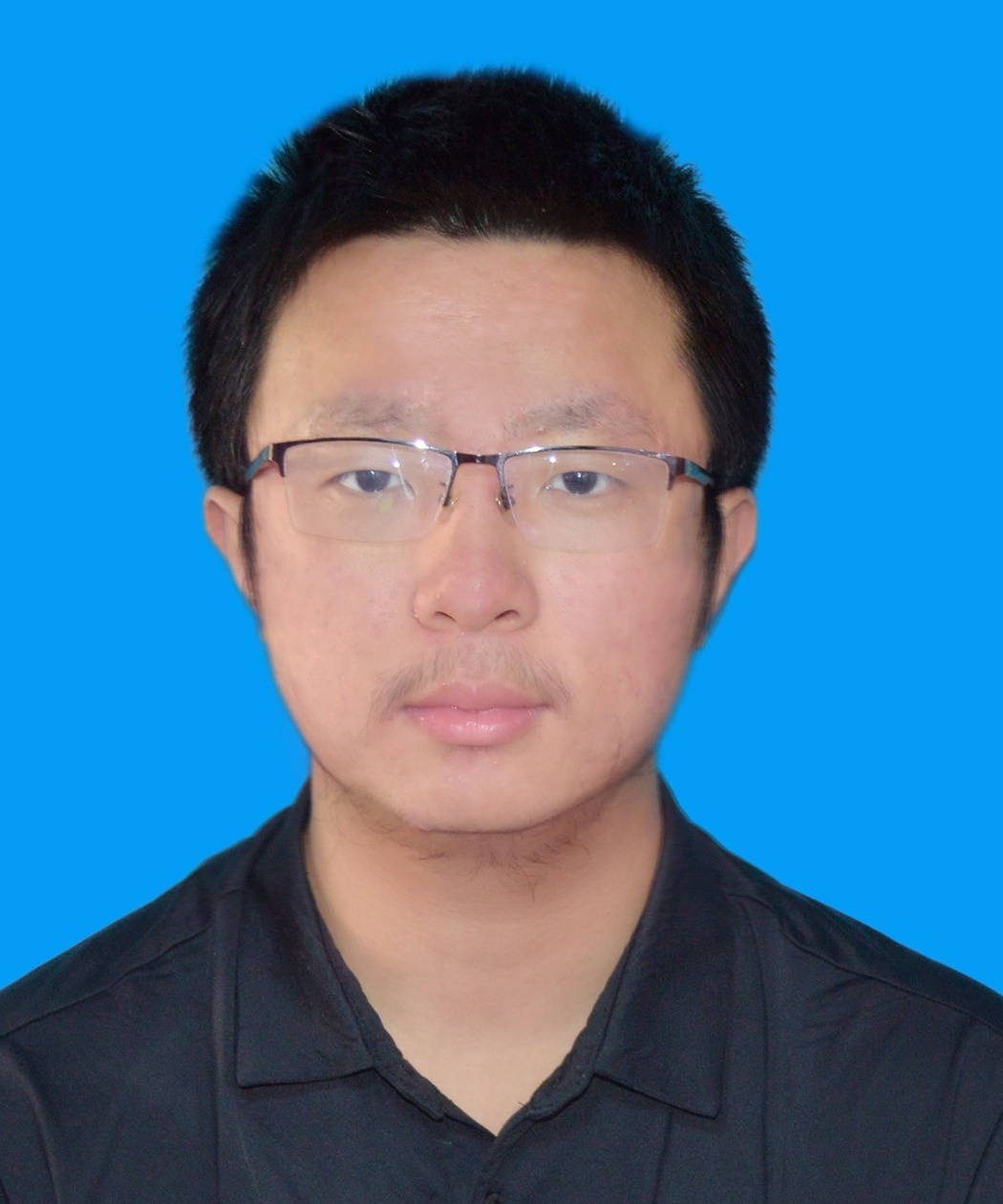}}]
{Taiyu Ban} received the B.Sc. degree in computer science and technology from the University of Science and Technology of China, Hefei, China, in 2020, where he is currently pursuing the Ph.D. degree in computer science and technology with the School of Computer Science and Technology.

His current research interests include causal discovery and causal-based machine learning.
\end{IEEEbiography}

\begin{IEEEbiography}[{\includegraphics[width=1in,height=1.25in,clip,keepaspectratio]{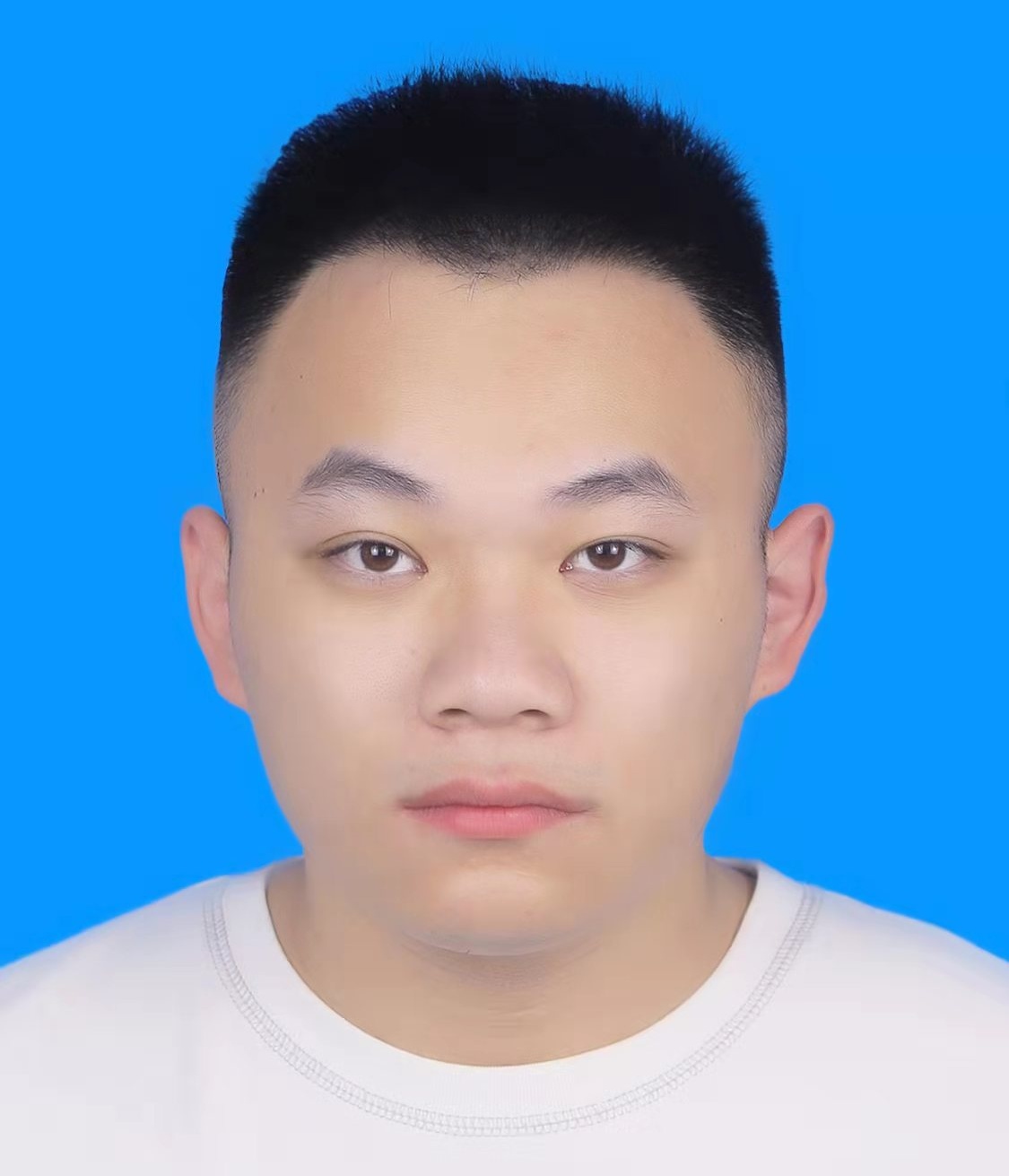}}]
{Lyuzhou Chen} received the B.Sc. degree in mathematics and applied mathematics from the University of Science and Technology of China, Hefei, China, in 2020, where he is currently pursuing the Ph.D. degree in computer science and technology with the School of Artificial Intelligence and Data Science.
  
His current research interests include causal discovery and causal-based machine learning.
\end{IEEEbiography}

\begin{IEEEbiography}[{\includegraphics[width=1in,height=1.25in,clip,keepaspectratio]{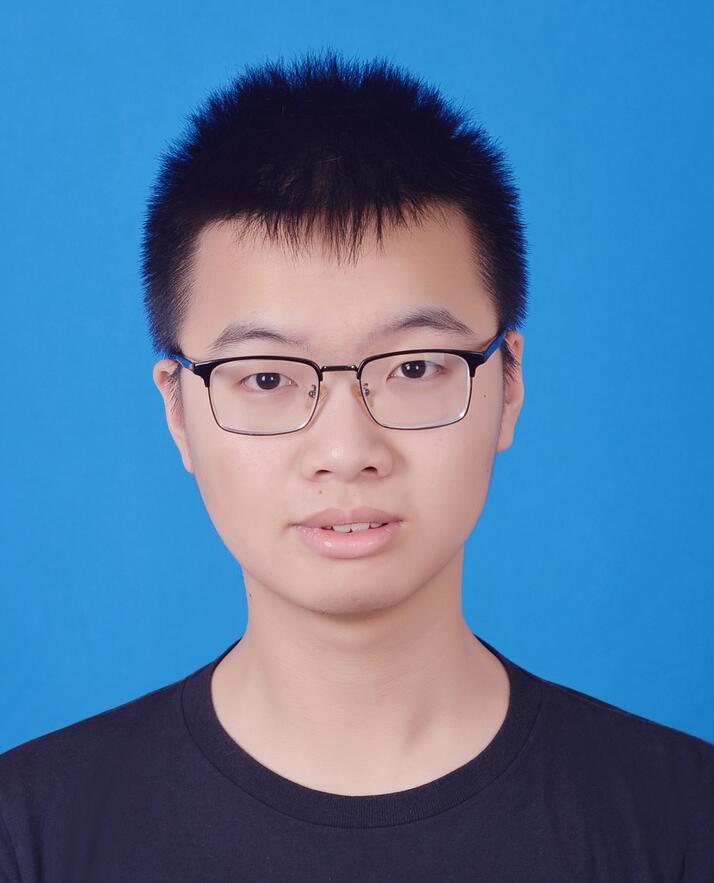}}]
{Derui Lyu} received the B.Sc. degree in intelligence science and technology from the School of Artificial Intelligence, Xidian University, Xi'an, China, in 2021. He is currently pursuing the Ph.D. degree in computer science and technology with the School of Computer Science and Technology, University of Science and Technology of China (USTC), Hefei, China.

His current research interests include machine learning and knowledge engineering.
\end{IEEEbiography}

\begin{IEEEbiography}[{\includegraphics[width=1in,height=1.25in,clip,keepaspectratio]{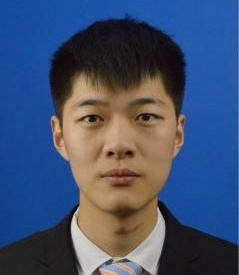}}]
{Xiangyu Wang} received the B.Sc. degree from the Donghua University, Shanghai, China, and the Ph.D. degree in data science from the University of Science and Technology of China (USTC). He is currently an associate researcher with the School of Computer Science and Technology, USTC. 

His current research interests include causal discovery and causal-based machine learning.
\end{IEEEbiography}

\begin{IEEEbiography}[{\includegraphics[width=1in,height=1.25in,clip,keepaspectratio]{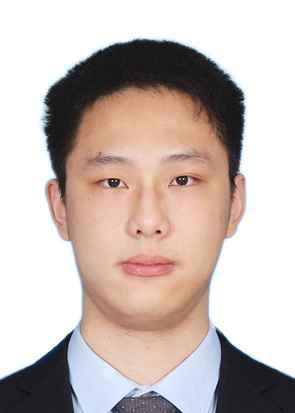}}]
{Qinrui Zhu} received the B.Sc. degree in computer science and technology from the University of Science and Technology of China, Hefei, China, in 2020, where he is currently pursuing the Ph.D. degree in computer science and technology with the School of Computer Science and Technology, and Laboratory for Big Data and Decision.

His research interests include large language models.
\end{IEEEbiography}

\begin{IEEEbiography}[{\includegraphics[width=1in,height=1.25in,clip,keepaspectratio]{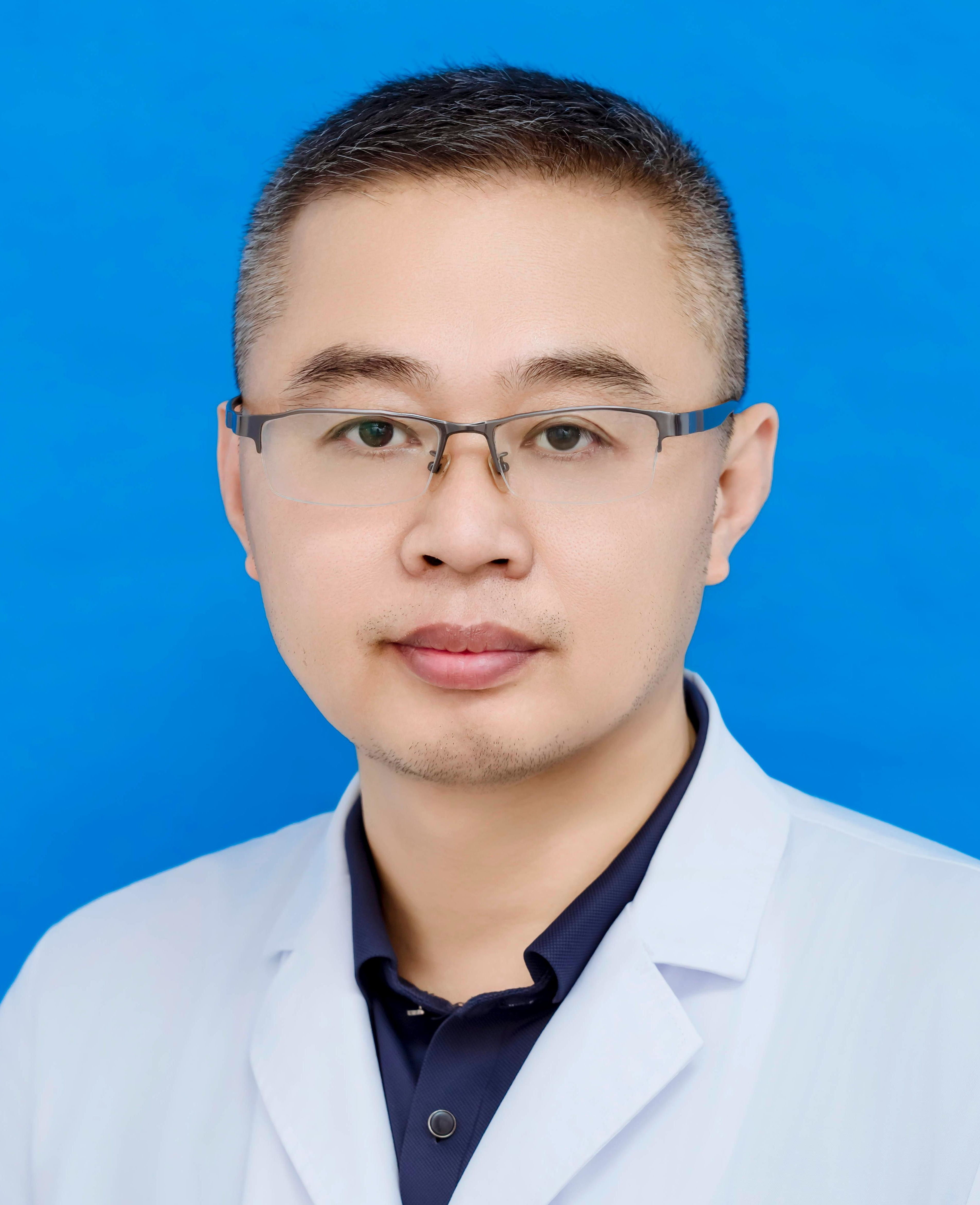}}]
{Qiang Tu} received the Ph.D. degree in computer science and technology from the University of Science and Technology of China (USTC). He is currently the director of the West District Information Center of the First Affiliated Hospital of USTC.

His research interests include machine learning and data mining.
\end{IEEEbiography}

\begin{IEEEbiography}[{\includegraphics[width=1in,height=1.25in,clip,keepaspectratio]{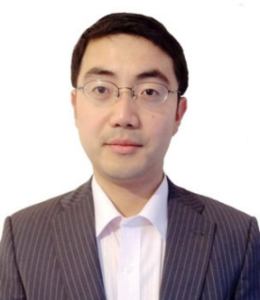}}]
{Huanhuan Chen} (Fellow, IEEE) received the B.Sc. degree from the University of Science and Technology of China (USTC), Hefei, China, and the Ph.D. degree in computer science from the University of Birmingham, Birmingham, U.K.
He is currently a Full Professor with the School of Computer Science and Technology, USTC. His current research interests include neural networks, Bayesian inference, and evolutionary computation.
Dr. Chen received the 2015 International Neural Network Society Young Investigator Award, the 2012 IEEE Computational Intelligence Society Outstanding Ph.D. Dissertation Award, the \textsc{IEEE Transactions on Neural Networks} Outstanding Paper Award (bestowed in 2011 and only one paper in 2009), and the 2009 British Computer Society Distinguished Dissertations Award.
He is now associate editor of the \textsc{IEEE Transactions on Neural Networks and Learning Systems}, and the \textsc{IEEE Transactions on Emerging Topics in Computational Intelligence}.
\end{IEEEbiography}

\end{document}